\documentclass[letterpaper]{article} 
\usepackage[preprint]{aaai2027}  
\usepackage[hyphens]{url}  
\usepackage{graphicx} 
\usepackage{natbib}  
\usepackage{caption} 
\usepackage{algorithm}
\usepackage{algorithmic}

\usepackage{newfloat}
\usepackage{listings}
\DeclareCaptionStyle{ruled}{labelfont=normalfont,labelsep=colon,strut=off} 
\floatstyle{ruled}
\newfloat{listing}{tb}{lst}{}
\floatname{listing}{Listing}

\usepackage{booktabs}

\usepackage{enumitem}
\usepackage{amsthm,amsmath,amsfonts,amssymb,stmaryrd}
\usepackage{mathtools}
\usepackage{pxfonts}

\usepackage{tikz}
\usetikzlibrary{automata, positioning, arrows,fit,patterns}

\usepackage{subcaption}

\usepackage[capitalize]{cleveref}
\usepackage{thm-restate}

\usepackage{xspace}

\usepackage{algorithm}
\usepackage{algorithmic}

\usepackage{pifont}
\newcommand{\cmark}{\ding{51}}
\newcommand{\xmark}{\ding{55}}
\usepackage{booktabs}
\usepackage{diagbox}

\newtheorem{theorem}{Theorem}
\newtheorem{proposition}[theorem]{Proposition}
\newtheorem{lemma}[theorem]{Lemma}
\newtheorem{corollary}[theorem]{Corollary}

\theoremstyle{definition}

\newtheorem{definition}[theorem]{Definition}
\newtheorem{example}[theorem]{Example}

\newcommand\pref[1]{Problem (\ref{#1})}

\newif\ifdraft
\draftfalse
\usepackage{import}
\usepackage{savesym}
\savesymbol{comment}

\ifdraft
\usepackage[draft]{commenting}
\else
\usepackage[nompar]{commenting}
\fi

\declareauthor{lo}{LO}{red} 
\authorcommand{lo}{comment}

\declareauthor{dw}{DW}{blue}
\authorcommand{dw}{comment}

\declareauthor{lxb}{LXB}{olive}
\authorcommand{lxb}{comment}

\declareauthor{lw}{LW}{purple}
\authorcommand{lw}{comment}

\declareauthor{ar}{AR}{magenta}
\authorcommand{ar}{comment}
\makeatletter
\renewcommand{\comm@todo@mpar}[1]{}
\makeatother

\def\divider{%
  \leavevmode\leaders\hrule height 0.6ex depth \dimexpr0.4pt-0.6ex\hfill%
  \kern0pt%
}

\renewcommand{\OpenCommentBraket}{$\lceil$}
\renewcommand{\ClosedCommentBraket}{$\rfloor$}

\newcommand\defeq\coloneqq
\newcommand\E{\mathbb{E}}

\newcommand\run\rho
\newcommand\runstate{\rho^S}
\newcommand\runaction{\rho^A}
\newcommand\lab\ell
\newcommand\tif{\text{if }}
\newcommand\tow{\text{otherwise}}
\newcommand\nat{\mathbb N}
\newcommand\real{\mathbb R}
\newcommand\ap{\mathcal{AP}}

\newcommand\mdp{\mathcal M}
\newcommand\mdpsub[1]{\mdp[#1]}

\newcommand\precc[3]{J_\mathrm{rec}(#1, #3, #2)}
\newcommand\preach[3]{\precc{#1}{#2}{#3}}

\newcommand\exavg[3]{J_\mathrm{avg}(#1, #3, #2)}

\DeclareMathOperator*{\clim}{C-lim}

\newcommand\mdpabs{\mathcal M_{\abs}}
\newcommand\pabs{\pi_{\abs}}

\newcommand\thresh{q}

\newcommand\mdpaug{\mathcal M_{F,F'}}
\newcommand\mc{\mdp}
\newcommand\policy\pi
\newcommand\pmem{\pi^{\mathrm{mem}}}

\newcommand\prob{\mathbb P}
\newcommand\dra{\mathcal A}
\newcommand\drac{\mathcal B}
\newcommand\rew{\mathcal R}

\newcommand\limavg[1]{\mathcal J_{#1^{\mathrm{avg}}}^\mdp}

\newcommand\lag\lambda

\newcommand\dist{\Delta}

\newcommand\alphab{\mathcal{AP}}

\newcommand\prodmdp{\mdp\otimes\dra\otimes\dra'}

\DeclareMathOperator{\infi}{Inf}

\DeclareMathOperator{\ing}{in}
\DeclareMathOperator{\out}{out}

\newcommand\distr{\dist_\policy^\mdp}
\newcommand\distrtwo[1]{\dist_{#1}^\mdp}
\newcommand\distrthree[2]{\dist_{#1}^{#2}}

\newcommand\ec{\mathcal C}
\newcommand\act{\mathrm{Act}}
\newcommand\mecs{\mathrm{MEC}}
\newcommand\pmecs{\mecs_p}
\newcommand\coll[1]{\widetilde{#1}}
\newcommand\collmdp[3]{\mathrm{Coll}(#1,#2,#3)}

\newcommand\ecr[1]{\ec(#1)}

\newcommand\stay{\textsc{stay}}

\DeclareMathOperator{\supp}{supp}

\newcommand\pmix{\overline\Pi}
\newcommand\ps{\Pi_{\mathrm{stat}}}
\newcommand\pd{\Pi^{\mathrm{det}}_{\mathrm{stat}}}
\newcommand\pmd{\Pi^{\mathrm{det}}_{\mathrm{mix}}}

\newcommand{\lift}[1]{{#1}_{\mathrm{lift}}}
\newcommand{\liftdet}[1]{{#1}_\mathrm{lift}^{\mathrm{det}}}

\newcommand\ind[1]{\mathbf 1_{\{#1\}}}

\title{Near-Optimal Reinforcement Learning for Constrained Recurrence Objectives}
\author{
Dominik Wagner\equalcontrib, Leon Witzman\equalcontrib, Luke Ong
}
\affiliations{
NTU Singapore\\
    \{dominik.wagner,luke.ong\}@ntu.edu.sg,  WITZ0001@e.ntu.edu.sg
}

\begin{document}

\maketitle

\begin{abstract}
Recurrence objectives, where a target region must be visited infinitely often, are a fundamental class of specifications for Markov decision processes (MDPs) and form the core of $\omega$-regular and linear temporal logic (LTL) objectives. We study constrained recurrence objectives, a natural extension of recurrence objectives with probabilistic constraints capable of modelling safety or fairness requirements. We first study the structure of optimal policies, showing that constrained recurrence requires different policy classes than those sufficient for other constrained MDP formalisms. In particular, we show that every feasible instance admits an optimal mixture of two stochastic stationary policies, as well as an optimal mixture of two deterministic stationary policies over a one-bit augmented MDP. We then study the generative-model reinforcement learning setting and propose an algorithm that first identifies the maximal end-component decomposition of the MDP, then reduces constrained recurrence to a constrained average reward problem for a collapsed MDP. Moreover, we establish a $\tilde{\mathcal{O}}(1/p+B/\varepsilon^2)$ sample complexity guarantee per state-action pair, where $p$ bounds certain non-zero transition probabilities and $B$ bounds transient time. Finally, we prove a nearly matching lower bound, showing that $1/p$ dependence, unlike the average-reward setting, is unavoidable.
\end{abstract}

\dw{
possible additional direction: strict feasibility setting
}

\dw{check $p$ vs. $\thresh$}

\dw{checklist}

\section{Introduction}

Markov decision processes (MDPs) provide a framework for
sequential decision making under uncertainty. Whilst much of the reinforcement
learning (RL) literature focuses on finite-horizon or discounted objectives,
many applications require guarantees over infinite executions. A naturally occurring requirement is \emph{recurrence}, which asks
that a designated set of states be visited infinitely often. Recurrence forms the foundation of linear temporal logic (LTL) and the moere general $\omega$-regular objectives via standard product constructions \citep{Bozkurt:2019,Perez2024,HPSS0W20}.

In this paper, we study constrained recurrence
problems in finite MDPs, which arise naturally when long-run performance must be balanced against
additional requirements. For example, an autonomous agent may seek to maximise
the probability of repeatedly servicing a collection of locations whilst
guaranteeing a minimum probability of repeatedly returning to a charging
station. Such formulations capture trade-offs between competing
long-run behaviours that cannot be expressed by unconstrained recurrence
alone. Moreover, constrained recurrence naturally
generalises several fundamental problems, including constrained reachability
and constrained reach-avoidance.

An alternative classical formalisation of infinite-horizon tasks uses average rewards \citep{Puterman:1994}, which quantifies the asymptotic rate of reward accumulation. In contrast, recurrence measures the probability that desirable events occur infinitely often, and presents new challenges for both policy class sufficiency and learning difficulty.

Whilst deterministic stationary policies suffice for unconstrained discounted, average-reward, and recurrence objectives \citep{Puterman:1994,Baier:2008}, constraints fundamentally alter the landscape. Constrained discounted problems admit optimal stochastic stationary policies \citep{Altman}, whereas stationary policies are insufficient for constrained average-reward and recurrence objectives. Moreover, unlike constrained average-reward problems, constrained recurrence problems do not always admit optimal mixtures of deterministic stationary policies (see \cref{ex:express}).

We establish that feasible constrained recurrence instances admit  optimal policies that are mixtures of two stochastic stationary policies, as well as mixtures of two deterministic stationary policies over a one-bit augmentation of the MDP. Our proof collapses maximal end components to reduce constrained recurrence to constrained reachability, and subsequently to constrained average reward, before lifting the resulting policies back to the original (or augmented) MDP.

We next turn to the RL setting, where finite sample guarantees require knowledge of structural parameters of the MDP beyond its size \citep{Alur:2022, Yang2022}.
The key algorithmic observation is that the preceding
reduction to constrained average reward requires only knowledge of the MDP's maximal end components, rather than its exact transition probabilities.
Hence, by first identifying the graph structure of the collapsed MDP with high
probability, we may then leverage recent generative-model RL algorithms for
constrained average-reward.

For that setting,
\citet{wei2025near} establish a sample complexity guarantee of
$\tilde{\mathcal O}((B+H)/\epsilon^2)$ per state-action pair, where $B$ bounds
the transient time before recurrent behaviour is reached and $H$ bounds the
bias span, measuring the variation in transient reward accumulated before
long-run average behaviour dominates \citep{zurek2024span}.
Whilst $B$ remains relevant for recurrence objectives, $H$ has no direct analogue since it is defined relative to a scalar
reward function. Notably, we show that \textit{$B$ alone is insufficient to obtain finite sample guarantees for recurrence objectives}.
Instead, we establish a $\tilde{\mathcal O}(1/p+B/\epsilon^2)$ sample complexity guarantee per state-action pair, where $p$ bounds certain non-zero transition probabilities, alongside a matching lower bound (up to logarithmic factors).

To the best of our knowledge, this constitutes the first near-optimal sample complexity guarantee for even \emph{unconstrained} recurrence problems in the generative-model setting.

\paragraph{Contributions.} Our contributions are threefold (see also \cref{tab:summary,tab:sample}). First, we characterise the structure of
optimal policies for constrained recurrence objectives. Second, we develop a $\tilde{\mathcal O}(1/p+B/\epsilon^2)$ generative-model RL algorithm, where $p$ bounds certain non-zero transition probabilities and $B$ is a transient time parameter. Third, we prove matching lower bounds (up to logarithmic
factors).

\paragraph{Related Work.}
The study of constrained MDPs \cite{Altman} and Safe RL (see \citep{gu2024review,WachiSS24} for surveys) typically involves problems defined by scalar reward and cost signals, which have limited ability to directly specify recurrence properties.

The statistical difficulty of RL has been studied in a variety of classical settings using generative models \citep{azar2012sample,zurek2024span,vaswani2022near,wei2025near}; see also \cref{tab:sample}.

\citet{ciesinski2008} study $\omega$-regular objectives (which extend recurrence) in the unconstrained planning setting, where the MDP is fully known. \citet{etessami2008multi,forejt2011quantitative} consider the multi-objective extension.

While $\omega$-regular specifications have received wide attention in RL (e.g.\ \citep{Wolff:12,V0CY22,Cai:2023,FuT:14}), most prior work does not establish sample complexity guarantees.

For the unconstrained problem, \citet{Perez2024} establish a (total) sample complexity guarantee of
$\tilde{\mathcal O}(|S|^3|A|T^3/\epsilon^4)$, with $T$ a mixing time parameter, using samples from on-policy trajectories
 rather than a generative model. They do not establish
lower bounds.

\cite{V0CY22,SVYVCS24} study the RL problem of minimising a scalar cost subject to satisfying an $\omega$-regular objective with maximal probability, while \cite{hahn2023multi} study multi-objective RL problems with lexicographic preference orderings. While distinct from the constrained recurrence problem studied in the present paper, both generalise the unconstrained recurrence setting. For the unconstrained problem, \cite{V0CY22} obtain a generative-model sample complexity of
$\tilde{\mathcal O}(1/p_{\min} + S^2B/\epsilon^4)$
per state-action pair, where $p_{\min}$ is a known lower bound on all positive transition probabilities. They do not establish lower bounds.

\begin{table}[t]
    \centering
    \begin{tabular}{lccccc}
\toprule
& \multicolumn{2}{c}{Stationary} & \multicolumn{2}{c}{Mix.\ Stat.} \\
& Det.
& Stoch.
& Det.
& Stoch. \\
\midrule
Unconstr. & \cmark & \cmark & \cmark & \cmark \\
\midrule
Disc. CMDP
& \xmark & \cmark & \cmark & \cmark & \\
\eqref{creach} & \xmark & \cmark$^\ddagger$ & \cmark & \cmark & \\
\eqref{cavg} & \xmark & \xmark & \cmark & \cmark & \\
\eqref{cra} & \xmark$^\dagger$ & \xmark$^\dagger$ & \cmark & \cmark\\
\eqref{crec}
& \xmark$^\dagger$ & \xmark$^\dagger$ & augm.$^\star$ & \cmark$^\star$ \\

\bottomrule
\end{tabular}
    \caption{ Sufficiency of policy classes. 
The first row summarises unconstrained discounted, average reward and recurrence problems. $^\ddagger$ is due to \citet{etessami2008multi}. $^\dagger$ is illustrated in \cref{ex:express}. \cref{ex:fig8} shows mixtures of stationary deterministic policies are insufficient for \eqref{crec}, whilst positive results $^\star$ are established in \cref{thm:type}. }
    \label{tab:summary}
\end{table}

\begin{table*}[t]
    \centering
    \begin{tabular}{ccccc}
\toprule
Uncon. Disc.\ reward
& Uncon. Avg.\ reward
& \multicolumn{2}{c}{Uncon. Reachability}
& Uncon. Recurrence \\
\midrule
$\tilde \Theta \left( \frac{1}{(1-\gamma)^3 \epsilon^2} \right)$ & $\tilde \Theta \left( \frac{H + B}{\epsilon^2} \right)$ & \multicolumn{2}{c}{$\tilde \Theta \left( \frac{B}{\epsilon^2} \right)$} & $\tilde \Theta \left( \frac{1}{p} + \frac{B}{\epsilon^2} \right)$ \\
\cite{azar2012sample} & \cite{zurek2024span} & \multicolumn{2}{c}{\cref{sec:reach}} & \cref{thm:recsample,cor:maincomplexity}
\\
\bottomrule
\\
\toprule
Disc.\ CMDP
& \eqref{cavg}
& \eqref{creach}
& \eqref{cra}
& \eqref{crec} \\
\midrule
$\tilde \Theta \left( \frac{1}{(1-\gamma)^3 \epsilon^2} \right)$ & $\tilde \Theta \left( \frac{H + B}{\epsilon^2} \right)$ & $\tilde \Theta \left( \frac{B}{\epsilon^2} \right)$ & $\tilde \Theta \left( \frac{B}{\epsilon^2} \right)$ & $\tilde \Theta \left( \frac{1}{p} + \frac{B}{\epsilon^2} \right)$ \\
\cite{vaswani2022near} & \cite{wei2025near} & \cref{sec:reach} & \cref{sec:reach} & \cref{thm:recsample,cor:maincomplexity}\\

\bottomrule
\end{tabular}
    \caption{Near-optimal generative-model RL sample complexity, per state-action pair. Parameters: optimality/feasibility gap $\epsilon$, discount factor $\gamma$, bias span bound $H$, transient time bound $B$, and internal transition probability bound $p$.}
    \label{tab:sample}
\end{table*}

\section{Background}
\label{sec:back}

For $N\in\nat$, let $[N]\defeq\{1,\ldots,N\}$, and let $\ind{\varphi}\in\{0,1\}$ denote the indicator of a predicate $\varphi$. For a countable set $X$, let $2^X$ denote its power set, $\dist(X)$ the set of probability distributions over $X$, and $\supp(\mu)\subseteq X$ the support of $\mu\in\dist(X)$.

We work with MDPs $\mdp=(S,A,\act, s_0,P)$, with $S$ and $A$ finite state and action sets, enabled actions $\act(s)\subseteq A$, initial state $s_0$, and transition kernel $P$. A state $s\in S$ is absorbing if for all $a\in\act(s)$, $P(s\mid s,a)=1$, and a set $S'\subseteq S$ is absorbing if every $s\in S'$ is absorbing. For $\mdp=(S,A,\act, s_0,P)$ and $s\in S$ we define $\mdp[s]\defeq(S,A,\act, s,P)$.

A \emph{run} $\run$ is an infinite sequence $\runstate_0\runaction_0\runstate_1\runaction_1\cdots\in (S\times A)^\omega$.

\paragraph{Policies.}
A \emph{history-based policy} is a function $\policy:(S\times A)^*\times S\to\dist(A)$ mapping histories $\runstate_0\runaction_0\cdots\runstate_n$ of past states and actions to a distribution $\dist(A)$ over next actions satisfying $\supp(\policy(\runstate_0\runaction_0\cdots\runstate_n))\subseteq\act(\runstate_n)$.
A history-based policy $\policy$ is \emph{stationary} if it only depends on the current state, i.e.\ $\policy(\runstate_0\cdots\runstate_m)=\policy((\run')^S_0\cdots(\run')^S_n)$ whenever $\runstate_m=(\run')^S_n$, and a history-based policy $\policy$ is deterministic if $\policy(\runstate_0\runaction_0\cdots\runstate_n)$ is a Dirac distribution for all $\runstate_0\runaction_0\cdots\runstate_n\in (S\times A)^*\times S$.

We use $\distr$ for the distribution over runs generated by a policy $\policy$,
and use $\ps(\mdp)$ and $\pd(\mdp)$ for the set of stationary and deterministic stationary policies, respectively.

For history-based policies $(\policy_i)_{i \in \nat}$ and non-negative coefficients $(\lambda_i)_{i \in \nat}$ with $\sum_{i \in \nat} \lambda_i = 1$, we call formal sums $\policy\defeq\sum_{i \in \nat} \lambda_i \cdot \policy_i$ \emph{mixture policies}, which generate the distribution $\distr\defeq\sum_{i \in \nat} \lambda_i \cdot \distrtwo{\policy_i}$ over runs. \lw{we can limit this to finite sums if needed} That is, a mixture policy randomly selects a (history-based) policy to follow for the full run.
Let $\pmix(\mdp)$ denote the set of mixture policies.

\paragraph{Generative Models.}
In the RL setting, the transition kernel $P$ is unknown.
Instead, algorithms may be given access to a \emph{generative model}, i.e.\ a
simulator that returns independent samples from $P(\cdot\mid s,a)$ for any
queried state-action pair $(s,a)\in S\times A$
\citep{kearns1998finite}.

\paragraph{Transient Time.}
For an MDP $\mdp$ and $\policy\in\ps(\mdp)$ let $R_\policy\subseteq S$ denote the set of recurrent states. For a run $\run$ and $S'\subseteq S$ let $\tau_{\run,S'}\defeq\inf\{t\in\nat\mid\runstate_t\in S'\}$ be the first hitting time of $S'$. 
Following \citet{zurek2024span}, for an MDP $\mdp$ we define the transient time $B_\mdp$
\begin{align*}
    B_\mdp\defeq\max_{\policy\in\pd}\max_{s\in S}\E_{\run\sim\distrthree\policy{\mdp[s]}}[\tau_{\run,R_\policy}]<\infty
\end{align*}

\paragraph{Average Reward Objectives.}
For $\mdp,\policy$ and a function $R:S\times A\to\real$, the expected average reward is defined by $\exavg\mdp\policy R\defeq \lim\inf_{T\to\infty}\E_{\run\sim\distr}\left[\frac 1 T\cdot\sum_{i=0}^{T-1} R(\runstate_i,\runaction_i)\right]$. 

The
\emph{bias-span} $H_{\mdp, R}$
 measures the variation in transient reward accumulated before long-run
average behaviour dominates \citep{zurek2024span}, \changed[dw]{see the appendix for details}.

\paragraph{Constrained Average Problems.}
For an MDP $\mdp$ and $R,R':S\times A\to[0,1]$, consider the optimisation problem
\begin{align}
\notag
    \max_{\pi\in\pmix(\mdp)}\; &\exavg\mdp\pi R\quad
\text{s.t.}\;
\exavg\mdp\pi{R'}\geq\thresh
\tag{C-Avg}
\label{cavg}
\end{align}
\citet{gonzalez2011optimal} show that feasible instances of \pref{cavg} admit optimal policies which are mixtures of deterministic stationary policies.

\paragraph{Near-Optimal Algorithm.}
 \citet{wei2025near}
 propose a randomised generative model RL algorithm \textsc{C-Avg} which, given an MDP without transition kernel $(S, A, \act, s_0)$, i.i.d.\ samples $\mathcal D=(d^{(i)}_{s,a,i})^{i\in[N]}_{s\in S,a\in A}$, reward/constraint functions $R,R':S\times A\to [0,1]$, threshold $\thresh\in[0,1]$, transient time bound $B$,  bias span bound $H$ and target accuracy $\epsilon\in(0,1)$, returns an $\epsilon$-optimal solution to feasible instances of \pref{cavg} with probability at least
\begin{align}
    1 - c\cdot |S||A|\cdot\exp\!\left(-c'\cdot{N\epsilon^2}/{(B+H)}\right)
\end{align}
 \citep[Thm.~2]{wei2025near}
where $c,c'>0$ are universal constants, provided that $H \geq \max(H_{\mdp,R}, H_{\mdp, R'})$ and $B \geq B_\mdp$. \dw{bias span, recall?}
In particular, the algorithm outputs mixtures of deterministic stationary policies.
This yields a $\tilde{\mathcal O}((B+H)/\epsilon^2)$ sample-complexity per state-action pair, alongside their nearly matching lower bound.

\section{Constrained Recurrence Problems}
\label{sec:setup}
We now define recurrence objectives and the corresponding constrained optimisation problem studied in
this paper.

For an MDP $\mdp=(S,A,\act,s_0,P)$, states $F\subseteq S$ and $\pi\in\pmix(\mdp)$, let
\begin{align*}
    \precc\mdp\pi F\defeq \prob_{\run\sim\distr}[\infi(\run)\cap F\neq\emptyset]
\end{align*} be the probability that a state in $F$ is visited infinitely often.\footnote{$\infi(\run)$ denotes the set of states occuring infinitely often in $\run$.}
In particular, if $F\subseteq S$ is absorbing, then $\precc\mdp\pi F$ is the probability to eventually reach a state in $F$. Likewise, if $U\subseteq S$ is absorbing then $\precc\mdp\pi{S\setminus U}$ is the probability to forever avoid all states in $U$.

\paragraph{Problem Statement.}
Given an MDP $\mdp$, sets $F,F'\subseteq S$ and a threshold $\thresh \in [0,1]$, the constrained recurrence problem is
\begin{align}
\notag
    \max_{\pi\in\pmix(\mdp)}\; &\precc\mdp\pi F\quad
\text{s.t.}\;
\precc\mdp\pi{F'}\ge\thresh
\tag{C-Rec}
\label{crec}
\end{align}
As is standard for RL, we consider \emph{$\epsilon$-optimal} policies $\policy\in\pmix(\mdp)$, which are feasible and optimal up to an $\epsilon$-margin:
\begin{enumerate}
    \item $\precc\mdp\pi {F'}\geq\thresh-\epsilon$ and
    \item $\precc\mdp\policy{F}\geq\precc\mdp{\policy'}F - \epsilon$ for all $\pi'\in\pmix(\mdp)$ satisfying $\precc\mdp{\policy'}{F'}\geq\thresh$.
\end{enumerate}

\paragraph{Special Cases: Reachability and Reach-Avoid.}
For absorbing $F,F',U\subseteq S$ we call the following spcecial cases \emph{constrained reachability} and \emph{constrained reach-avoid}, respectively:
\begin{align}
    \max_{\pi\in\pmix(\mdp)}\; &\precc\mdp\pi F\quad
\text{s.t.}\;
\precc\mdp\pi{F'}\ge\thresh
\tag{C-Reach}
\label{creach}\\
    \max_{\pi\in\pmix(\mdp)}\; &\precc\mdp\pi F\quad
\text{s.t.}\;
\precc\mdp\pi{S\setminus U}\ge\thresh
\tag{C-RA}
\label{cra}
\end{align}
Unlike the classical unconstrained reach-avoid objective, \pref{cra} decouples reaching $F$
from avoiding $U$; the latter is enforced as a probabilistic constraint.

\begin{figure}
    \begin{center}
    \begin{minipage}[b]{0.55\linewidth}
    \begin{subfigure}{\linewidth}
    \begin{center}
    \scalebox{0.6}{
            \begin{tikzpicture}[node distance = 2cm]
            \node[state, accepting, very thick, color = green!60!black, initial] (s0) {$s_0$};
            \node[state, accepting, very thick, color = green!60!black, below left of=s0] (s1) {$s_1$};
            \node[state,accepting,minimum size=0.7cm] at (s1.center) {};
            \node[state, below right of=s0] (s2) {$s_2$};
              \draw[->] 
              (s0) edge [loop right] node [right]{$a/1$} (s0)
                (s0) edge node [below right,xshift=-4pt]{$b/\frac{1}{2}$} (s1)
                (s0) edge node [below left,xshift=4pt]{$b/\frac{1}{2}$} (s2)
                (s1) edge [loop left] node [left]{$a/1,b/1$} (s1)
                (s2) edge [loop right] node [right]{$a/1,b/1$} (s2);
        \end{tikzpicture}    
        }
        \end{center}
    \caption{MDP for \cref{ex:express}.}
    \label{fig:rps}        
    \end{subfigure}

    \begin{subfigure}{\linewidth}
    \begin{center}
    \scalebox{0.6}{
        \begin{tikzpicture}[node distance = 2cm]
            \node[state, initial above] (v1) {$s_0$};
            \node[state, accepting,  left of=v1] (v4) {$s_1$};
            \node[state, accepting, color=green!60!black, very thick, right of=v1] (v5) {$s_2$};

              \draw[->] 
              (v1) edge [bend right] node [above]{$a$} (v4)
              (v4) edge [bend right] node [right=2pt, above]{$a,b$} (v1)
              (v1) edge [bend left] node [above]{$b$} (v5)
              (v5) edge [bend left] node [left=2pt, above]{$a,b$} (v1)
            
              ;
        \end{tikzpicture}     
        }
    \end{center}
    \caption{MDP for \cref{ex:fig8}.}
    \label{fig:eight}
\end{subfigure}
    \end{minipage}
    \hfill
    \begin{minipage}[b]{0.4\linewidth}
        \begin{subfigure}{\linewidth}
    \begin{center}
    \scalebox{0.6}{
        \begin{tikzpicture}[node distance = 2cm]
            \node[state, initial above] (s00) {$s_0,0$};
            \node[state, accepting, left of=s00] (s10) {$s_1,0$};
            \node[state, below of=s00] (s01) {$s_0,1$};
            \node[state, accepting, color=green!60!black, very thick, right of=s01] (s21) {$s_2,1$};
              \draw[->] 
              (s00) edge [bend right] node [above]{$a$} (s10)
              (s10) edge [bend right] node [above]{$a,b$} (s00)
              (s00) edge node [above right]{$b$} (s21)
              (s21) edge [bend left] node [below]{$a,b$} (s01)
              (s01) edge [bend left] node [above]{$b$} (s21)
              (s01) edge node [below left]{$a$} (s10);
        \end{tikzpicture}    
        }
    \end{center}
    \caption{Augmented MDP for \cref{fig:eight} (unreachable states are omitted).}
    \label{fig:eightaug}
\end{subfigure}
    \end{minipage}
    \end{center}
    \caption{MDPs illustrating insufficient policy classes for \pref{crec}.}
\end{figure}

\paragraph{Insufficient Policy Classes.}
Whilst stochastic stationary policies suffice for discounted CMDPs \citep{Altman}, the following examples show that they are insufficient for
constrained recurrence.
\begin{example}
\label{ex:express}
    The MDP in \cref{fig:rps} is an instance of \pref{cra} with target $F\defeq \{s_1\}$, avoid-set $U\defeq \{s_2\}$, and threshold $\thresh \defeq 0.9$. Clearly,
    $\precc M \policy F = 1 - \precc M \policy { S \setminus U } $ for all policies $\policy$. Thus, any policy which (eventually) leaves $s_0$ with probability exactly $0.2$ would be a solution. No deterministic nor stationary policy can achieve this, as any such policy either remains in $s_0$ almost surely or leaves $s_0$ almost surely. It is however achieved by the mixture $0.8 \cdot \pi_a + 0.2 \cdot \pi_b$, where $\pi_a$ is the deterministic policy always performing action $a$, likewise for $b$.
\end{example}
Furthermore, mixtures of deterministic stationary policies (a sufficient class for constrained average reward problems, \citep{gonzalez2011optimal}) are still insufficient to solve \pref{crec} in general.
\begin{example}
\label{ex:fig8}
    Consider the MDP in \cref{fig:eight}, with $F\defeq\{s_1\}$ and $F'\defeq\{s_2\}$. A deterministic stationary policy visits exactly one of $F, F'$ infinitely often in all runs. Therefore, for $\thresh>0$, any feasible mixture of deterministic stationary policies satisfies the objective with probability less than $1$, whereas alternating between $a$ and $b$ in $s_0$ satisfies both objective and constraint with probability $1$.
\end{example}

\paragraph{Augmented Recurrence Problem.}
Intuitively, the need to oscillate between two sets $F$ and $F'$ can preclude deterministic stationary policies. To address this limitation, we consider an \emph{augmented MDP} $\mdpaug$ with states $S \times \{0,1\}$, where the additional bit is updated deterministically to $0$ when visiting $F$, and to $1$ for $F'\setminus F$ (see \cref{def:mdpaug} in the appendix and \cref{fig:eightaug}).
Together with the augmented target sets
$F\times\{0,1\}$ and $F'\times\{0,1\}$, this defines the corresponding
\emph{augmented recurrence problem}.

\section{Warm-Up: Constrained Reachability}
\label{sec:reach}
We first study constrained reachability, which underpins the policy and sample-complexity results developed subsequently for constrained recurrence.

The key observation
is that for absorbing target sets $F\subseteq S$, reachability probabilities coincide
with average rewards:
\begin{align}
\label{eq:reachavg}
\preach{\mdp}{\policy}{F}=\exavg{\mdp}{\policy}{R_F}\quad\text{for } R_F(s,a)\defeq\ind{s\in F}
\end{align}
Consequently, constrained reachability reduces directly to constrained
average-reward. We therefore inherit both sufficient classes of optimal policies and
algorithmic results from the average-reward setting.

\paragraph{Optimal Policies.}
Combining \cref{eq:reachavg} with the policy characterisation of constrained
average-reward optimisation~\citep{gonzalez2011optimal} yields the
following.

\begin{proposition}
\label{prop:policyreach}
Every feasible instance of \pref{creach} admits an optimal policy that is a
mixture of two deterministic stationary policies.
\end{proposition}

\paragraph{Generative-Model Reinforcement Learning.}
By the above reduction, it
suffices to apply the average-reward algorithm \textsc{C-Avg}, see \cref{sec:back}. The remaining
ingredient is a bound on the bias span, which for reachability rewards $R_F$ is
controlled by the transient time.
\begin{restatable}{lemma}{spantrans}
\label{lem:spantrans}
Let $\mdp=(S,A,\act,s_0,P)$ be an MDP, for which $F\subseteq S$ is absorbing. Then for $R_F(s,a)\defeq\ind{s\in F}$, $H_{\mdp,R_F}\leq B_\mdp$.
\end{restatable}

Combining \cref{eq:reachavg,lem:spantrans} with
\citep[Theorem~2]{wei2025near} yields the following sample
complexity guarantee.

\begin{proposition}
\label{prop:reachrl}
    Let $\mdp=(S,A,\act,s_0,P)$ be an MDP satisfying $B_\mdp\leq B$, let $F,F'\subseteq S$ be absorbing, $\thresh\in[0,1]$ and $\epsilon\in(0,1)$. If \pref{creach} is feasible then with probability at least
\begin{align*}
    1 - c\cdot|S||A|\cdot\exp\!\left(-c'\cdot{N\epsilon^2}/{B}\right)
\end{align*}
\cref{alg:rlreach}
 returns an $\epsilon$-optimal solution to \pref{creach},
where $c,c'>0$ are universal constants. 
Moreover, it returns a mixture of stationary deterministic policies.
\end{proposition}
Ignoring logarithmic factors, this corresponds to a sample complexity of
$\tilde{\mathcal O}(B/\epsilon^2)$ per state-action pair.

\paragraph{Lower Bound.}
\citet[Thm.~4]{zurek2024span} give a class of unconstrained average-reward problems demonstrating an $\Omega(B/\epsilon^2)$ generative model sample requirement per state-action pair.
With minor modifications, each instance in their class naturally translates to an equivalent unconstrained reachability problem while asymptotically preserving the relevant parameters $|S|,|A|,\epsilon,\delta,B$ (see appendix). Therefore, \cref{alg:rlreach} is optimal up to logarithmic factors.

\paragraph{Constrained Reach-Avoid.}
For absorbing $U\subseteq S$, the probability of always avoiding $U$ is
\begin{align*}
    \precc\mdp\policy{S\setminus U}=
\exavg\mdp\policy{R_{S\setminus U}}
\end{align*}
for $R_{S\setminus U}(s,a)\defeq\ind{s\in S\setminus U}$.
Consequently, constrained reach-avoid problems can likewise be reduced to constrained average reward. Hence, they admit optimal mixtures of deterministic stationary policies, and a straightforward modification of \cref{alg:rlreach} yields a sample complexity of $\tilde{\mathcal O}(B/\epsilon^2)$ per state-action pair. Together with the lower bound of \cref{thm:lowerboundb}, this is optimal up to logarithmic factors.

\begin{algorithm}[t]
    \caption{\textsc{C-Reach}}
    \label{alg}
    \begin{algorithmic}[1] 
    \REQUIRE MDP without transition kernel $(S, A, \act, s_0)$, i.i.d.\ samples $\mathcal D=(d^{(i)}_{s,a,i})^{i\in[N]}_{s\in S,a\in A}$, recurrence sets $F,F'\subseteq S$, threshold $\thresh\in[0,1]$, transient time bound $B$ and target accuracy $\epsilon\in(0,1)$
    
    \STATE Define $R_F$ and $R_{F'}$ as in \cref{eq:reachavg}
    \RETURN \textsc{C-Avg}$((S, A, \act, s_0), \mathcal D,R_F,R_{F'},p, B,B,\epsilon)$\hfill 
    \end{algorithmic}
\label{alg:rlreach}
\end{algorithm}

\section{Reducing Constrained Recurrence to Constrained Reachability}
\label{sec:LP}
We now extend the reachability results of \cref{sec:reach} to constrained
recurrence. We first establish a reduction to constrained reachability, from
which we derive sufficient classes of optimal policies. The reduction will
also form the basis of our generative-model RL results in the next section.

The reduction exploits maximal end components (MECs). Almost surely, the
state-action pairs visited infinitely often by a run form an end component, which is contained in a unique MEC. Consequently, a run
can satisfy a recurrence objective only if this MEC intersects the
corresponding target set. Conversely, once such a MEC has been reached,
selecting enabled actions uniformly at random visits every state of the MEC
infinitely often almost surely. It therefore suffices to reach an appropriate
MEC.

\paragraph{Maximal End Components.}
An \emph{end component} (EC) of an MDP $\mdp=(S,A,\act,s_0,P)$ is a pair
$(S_\ec,\act_\ec)$, where $S_\ec\subseteq S$ and
$\act_\ec:S_\ec\to 2^A$, satisfying:
\begin{enumerate}
    \item $\act_\ec(s)\subseteq\act(s)$ for every $s\in S_\ec$,
    \item $\sum_{s'\in S_\ec}P(s'\mid s,a)=1$ for every
    $s\in S_\ec$ and $a\in\act_\ec(s)$,
    \item the graph $(S_\ec,\rightarrow_{\act_\ec})$ is strongly connected,
    where $s\rightarrow_{\act_\ec}s'$ iff
    $P(s'\mid s,a)>0$ for some $a\in\act_\ec(s)$.
\end{enumerate}
We permit singleton ECs $({s},\act_\ec)$ with $\act_\ec(s)=\emptyset$ and call all other ECs \emph{proper}.

An EC $\ec$ is \emph{contained} in an EC $\ec'$ if
$S_\ec\subseteq S_{\ec'}$ and
$\act_\ec(s)\subseteq\act_{\ec'}(s)$ for every $s\in S_\ec$.
A \emph{maximal end component (MEC)} is an EC maximal under this partial order.

Distinct MECs have disjoint state sets and therefore partition the state
space $S$ (see \cref{fig:DAG}). We write $\mecs(\mdp)$ and $\pmecs(\mdp)$
for the sets of MECs and proper MECs of $\mdp$, respectively.
 Finally, for an EC
$\ec=(S_\ec,\act_\ec)$, we define
\begin{align*}
    \out(\ec)
\defeq
\{(s,a)\in S_\ec\times A
\mid
a\in\act(s)\setminus\act_\ec(s)\}
\end{align*}
that is, the state-action pairs leaving $\ec$ in a single step.

\begin{figure}[t]
    \begin{center}
    \begin{subfigure}{0.55\linewidth}
            \scalebox{0.45}{
        \begin{tikzpicture}[node distance = 2cm]
            \node[state, initial] (v0) {$s_0$}; 
            \node[state, below=15mm of v0] (v2) {$s_2$};
            \node[state, left=23mm of v2] (v1) {$s_1$};
            \node[state, accepting, color=green!60!black, very thick, right=23mm of v2] (v3) {$s_3$};
            \node[state, accepting, below left of=v1] (v4) {$s_4$};
            \node[state, accepting, color=green!60!black, very thick, below right of=v1] (v5) {$s_5$};
            \node[state, accepting, below of=v3] (v6) {$s_6$};

            \node[below=7mm of v2] (aux1) {};
            \node[right=7mm of v3] (aux2) {};
            \node[right=7mm of v6] (aux3) {};

            \node[draw, dashed, thick, fit=(v0), inner sep=10pt, label={[anchor=north west]north west:$\ec_0$}] (dashedbox) {};
            \node[draw, dashed, thick, fit=(v1) (v4) (v5), inner sep=10pt, label={[anchor=north west]north west:$\ec_1$}] (dashedbox) {};
            \node[draw, dashed, thick, fit=(v2) (aux1), inner sep=10pt,label={[anchor=north west]north west:$\ec_2$}] (dashedbox) {};
            \node[draw, dashed, thick, fit=(v3) (aux2), inner sep=10pt,label={[anchor=north west]north west:$\ec_3$}] (dashedbox) {};
            \node[draw, dashed, thick, fit=(v6) (aux3), inner sep=10pt,label={[anchor=north west]north west:$\ec_4$}] (dashedbox) {};
            
              \draw[->] 
              (v0) edge [loop above] node [above]{$a/0.2$} (v0)
              (v0) edge node [left=3pt, pos = 0.3]{$a/0.6$} (v1)
              (v0) edge node [left]{$a/0.2$} (v2)
              (v0) edge node [right=3pt, pos=0.3]{$b$} (v3)
              (v1) edge [bend right] node [above]{$a$} (v4)
              (v4) edge [bend right] node [right=2pt, pos=0.2]{$a,b$} (v1)
              (v1) edge [bend left] node [above]{$b$} (v5)
              (v5) edge [bend left] node [left=2pt, pos=0.2]{$b$} (v1)
              (v5) edge [bend right=40] node [below]{$a$} (v3)
              (v2) edge [loop below] node [below]{$a,b$} (v2)
              (v3) edge [loop right] node [right]{$a$} (v2)
              (v3) edge  node [right]{$b$} (v6)
              (v6) edge [loop right] node [right]{$a,b$} (v6)

              ;
        \end{tikzpicture} 
        }
    \caption{MDP with decomposition into MECs (dashed).}
    \label{fig:DAG}
    \end{subfigure}
    \hfill
    \begin{subfigure}{0.4\linewidth}
            \scalebox{0.47}{
        \begin{tikzpicture}[node distance = 2cm]
            \node[state, initial] (v0) {$\ec_0$}; 
            \node[state, below=15mm of v0] (v2) {$\ec_2$};
            \node[left=23mm of v2] (aux) {};
            \node[state, left=10mm of v2,yshift=-5mm] (v1) {$\ec_1$};

            \node[state, right=23mm of v2, yshift=10mm] (v3) {$\ec_3$};
            \node[state, below of=v3] (v6) {$\ec_4$};
            \node[state, below=15mm of v2] (bot) {$\bot$};
            \node[state, accepting, color=green!60!black, very thick, left=10mm of bot] (botFFp) {$\bot_{F,F'}$};
            \node[state,accepting,minimum size=0.87cm] at (botFFp.center) {};

            \node[state, accepting, color=green!60!black, very thick, right=10mm of bot] (botFp) {$\bot_{F'}$};
            \node[state, accepting, right=10mm of botFp] (botF) {$\bot_{F}$};

              \draw[->] 
              (v0) edge [loop above] node [above]{$(s_0,a)/0.2$} (v0)
              (v0) edge node [left]{$(s_0,a)/0.6$} (v1)
              (v0) edge node [right]{$(s_0,a)/0.2$} (v2)
              (v0) edge node [right=3pt, pos=0.3]{$(s_0,b)$} (v3)
              (v1) edge [bend right=40] node [pos=0.2,above]{$(s_5,a)$} (v3)
              (v3) edge  node [right]{$(s_3,b)$} (v6)
              (v2) edge node [right] {$\stay$} (bot)
              (v3) edge node [left] {$\stay$} (botFp)
              (v6) edge node [right] {$\stay$} (botF)
              (v1) edge node [left] {$\stay$} (botFFp)
              ;
        \end{tikzpicture} 
        }
    \caption{Collapsed MDP.}
    \label{fig:abs}
    \end{subfigure}
    \end{center}
\caption{Example for reduction ($F\defeq\{s_4,s_6\}, F'\defeq\{s_3,s_5\}$). Probabilities which are $1$ are omitted.}
\label{fig:running}
\end{figure}

\paragraph{Collapsed MDP.}
Following \citet{ciesinski2008,forejt2011quantitative}, we collapse every
maximal end component (MEC) into a single state. Intuitively, once a MEC has
been reached, the only remaining decision is whether to leave it via one of
its exit state-action pairs or to commit to remaining in the MEC forever. 
The collapsed MDP therefore has one state per MEC together with absorbing
states $\{\bot,\bot_F,\bot_{F'},\bot_{F\land F'}\}$ recording which recurrence objectives are satisfied by committing to
the current MEC.
State-action pairs $(s,a)\in\out(\ec)$ constitute enabled \emph{actions} of MECs $\ec$, while proper MECs
additionally admit a distinguished action $\stay$ representing the decision to
remain in $\ec$ indefinitely.

The transition kernel is defined by aggregating transition probabilities
between maximal end components. For every
$\ec,\ec'\in\mecs(\mdp)$ and $(s,a)\in\out(\ec)$,
\begin{align}
\label{eq:defcollp}
\coll P(\ec'\mid\ec,(s,a))
\defeq
\sum_{s'\in S_{\ec'}}P(s'\mid s,a)
\end{align}
Thus, transitions between collapsed states are obtained by aggregating the
probability mass of successors belonging to the same MEC. Moreover, executing
$\stay$ deterministically transitions to one of four absorbing states
according to whether the current MEC intersects neither, only $F$, only $F'$,
or both target sets. Since every state of a proper MEC can be visited
infinitely often almost surely, the absorbing state records precisely which
recurrence objectives are satisfied by committing to the MEC. In contrast to
the unconstrained setting, distinguishing these four outcomes is essential for
capturing the constrained recurrence objective. The full definition
is deferred to the appendix.

\paragraph{Policy Lifting.}
For the remainder of this section, fix an MDP
$\mdp=(S,A,\act,s_0,P)$, target sets $F,F'\subseteq S$, and let
$\coll\mdp\defeq\collmdp{\mdp}{F}{F'}$.

For every deterministic stationary policy
$\policy\in\pd(\coll\mdp)$, we define its \emph{lift}
$\lift\policy\in\ps(\mdp)$ as follows. Suppose $s\in S$ and let
$\ec=(S_\ec,\act_\ec)$ denote the unique MEC containing $s$.
If $\policy(\ec)=(s,a)\in\out(\ec)$, then
$\lift\policy$ selects action $a$ with probability~$1$ in state~$s$.
Otherwise, $\lift\policy$ chooses uniformly from $\act_\ec(s)$.
Thus, $\lift\policy$ follows the exit decisions prescribed by
$\policy$ and otherwise remains within the current MEC.

Similarly, every $\policy\in\pd(\coll\mdp)$ admits a deterministic lift
$\liftdet\policy\in\pd(\mdpaug)$ to the augmented MDP
(\cref{def:mdpaug}). Intuitively, $\liftdet\policy$ replaces the
stochastic behaviour within MECs by deterministic navigation: it follows fixed
paths to prescribed exit states and, after committing to remain in a MEC,
uses the augmentation bit to alternate between $F$ and $F'$ whenever
necessary, see the appendix for details.

\begin{restatable}{lemma}{absone}
\label{lem:abs1}
    For $\policy\in\pd(\coll\mdp)$ it holds \changed[dw]{(analogously for $F'$)}
    \begin{align*}
        \precc\mdp{\lift\policy}F&=\preach{\coll\mdp}{\policy}{\{\bot_F,\bot_{F\land F'}\}}\\
        \precc{\mdpaug}{\liftdet\policy}{F\times\{0,1\}}&=\preach{\coll\mdp}{\policy}{\{\bot_F,\bot_{F\land F'}\}}
    \end{align*}
\end{restatable}
\changed[dw]{NB The right-hand sides are reachability probabilities since the target states of the collapsed MDP are absorbing.}

\paragraph{Projecting Policies to the Collapsed MDP.}
We now establish the converse direction. Unlike the lifting construction, we
begin with an arbitrary policy of the original MDP, which may be
history-dependent, and construct a stationary policy for the collapsed MDP.

The key observation is that $\coll\mdp$ is absorbing, so
every policy induces finite occupancy measures, that is, finite expected
visitation counts for state-action pairs (see appendix).

Given a policy $\policy$ for $\mdp$, we construct an occupancy measure on
$\coll\mdp$. For every exit action $(s,a)\in\out(\ec)$, the corresponding
occupancy variable is the expected number of times $(s,a)$ is executed under
$\policy$. The occupancy variable of the distinguished action $\stay$ in a
proper MEC $\ec$ is the probability that $\policy$ eventually remains in
$\ec$ forever. We show that these quantities satisfy the occupancy-measure
constraints of $\coll\mdp$. The standard correspondence between feasible
occupancy measures and stationary policies for absorbing MDPs then yields the
desired policy.
\begin{restatable}{lemma}{proj}
\label{lem:proj}
    For $\policy\in\pmix(\mdp)$ there exists $\coll\policy\in\ps(\coll\mdp)$ s.t.\
    \begin{align*}
        \preach{\coll\mdp}{\coll\policy}{\{\bot_F,\bot_{F\land F'}\}}&\geq\precc\mdp\policy F\\
        \preach{\coll\mdp}{\coll\policy}{\{\bot_{F'},\bot_{F\land F'}\}}&\geq\precc\mdp\policy {F'}
    \end{align*}
\end{restatable}

\paragraph{Sufficient Policy Classes.}
Combining \cref{lem:proj,prop:policyreach,lem:abs1}, and applying the lifting
construction componentwise to the two deterministic policies in the optimal
mixture guaranteed by \cref{prop:policyreach}, yields the
following sufficient policy classes for \pref{crec}.

\begin{restatable}[Sufficient Policy Classes]{theorem}{typepolicy}
\label{thm:type}
For every feasible instance of \pref{crec},
\begin{enumerate}
    \item there exists an optimal policy that is a mixture of two stochastic
    stationary policies;
    \item the corresponding instance of the augmented recurrence problem
    admits an optimal policy that is a mixture of two deterministic stationary
    policies.
\end{enumerate}
\end{restatable}

\section{Generative-Model Reinforcement Learning for Constrained Recurrence}
\label{sec:rl}
Next, we propose a reinforcement learning algorithm for \pref{crec} which
receives
$N$ i.i.d.\ sample $d^{(i)}_{s,a}\sim P(\cdot\mid s,a)$ per state-action
pair. A natural approach is to reduce constrained recurrence to constrained
reachability via the collapsed MDP and then apply
\cref{alg:rlreach}. The difficulty is that the collapsed MDP depends on the
unknown transition kernel of the original MDP.

The key observation is that \textit{the reduction neither requires the full transition
kernel nor even the full support relation}. Rather, it suffices to identify the
MECs: once $\mecs(\mdp)$ is known, every aspect of the collapsed MDP
$\collmdp{\mdp}{F}{F'}$ is uniquely determined aside from its transition
kernel. 

Since MECs depend only on the support
relation
\begin{align*}
    \supp(P) \defeq \{ (s,a,s') \in S \times A \times S \mid P(s' \mid s,a) > 0 \}
\end{align*} we may equivalently write
$\mecs(S,A,\act,E)$ for the MECs induced by a support
relation $E\subseteq S\times A\times S$, so that
\[
\mecs(S,A,\act,\supp(P))=\mecs(\mdp).
\]

\begin{algorithm}[t]
    \caption{Reinforcement Learning for \pref{crec}}
    \label{alg:rlrec}
    \begin{algorithmic}[1] 
    \REQUIRE MDP without transition kernel $(S,A,\act,s_0)$, i.i.d.\ samples $\mathcal D=(d^{(i)}_{s,a})^{i\in[N]}_{s\in S,a\in A}$, regions $F,F' \subseteq S$, threshold $\thresh \in [0,1]$, transient time bound $B$ and target accuracy $\epsilon\in(0,1)$
    \STATE Construct empirical support $\widehat E$ as in \cref{eq:estsupp}
    \STATE Construct the collapsed structure $(\coll S,\coll A,\coll\act,\coll s_0)$ for $(S,A,\act,s_0)$ and $\widehat E$
    \STATE Construct the collapsed dataset $\coll {\mathcal D}$ from $\mathcal D$ using \cref{eq:collsamples}
    \STATE $\pi\gets$ \textsc{C-Reach}$\big(\left(\coll S,\coll A,\coll\act,\coll s_0\right), \coll {\mathcal D},$ \label{alg:creachline}
    
    \hfill $\{\bot_{F},\bot_{F\land F'}\}, \{\bot_{F'},\bot_{F\land F'}\}, p, B+1,\epsilon\big)$ 
    \RETURN $\lift\pi$ \COMMENT{lift policy to $\mdp$}
    \end{algorithmic}
\label{alg:rl}
\end{algorithm}

\paragraph{Recovering MECs.}
Our algorithm therefore begins by constructing the empirical support relation
$\widehat E$. 
\begin{align}
\label{eq:estsupp}
    \widehat E\defeq\{(s,a,s')\in S\times A\times S\mid\exists i\in[N]\text{ s.t.\ } d^{(i)}_{s,a}=s'\}
\end{align}
Whilst $\widehat E$ may underapproximate the true support, it suffices that
it induces the true MECs with high probability. To quantify
this probability, we require not only the transient-time parameter $B_\mdp$,
but also the following structural parameter.
\begin{definition}[$p_\mdp$ Parameter]
    For $\mdp = (S,A,\act,s_0,P)$, we call a transition $(s,a,s') \in S \times A \times S$ \emph{internal} when $s$ and $s'$ belong to the same MEC $\ec$ and $a \in \act_\ec(s)$. Define $p_\mdp$ as the smallest non-zero
probability of an internal transition.
\end{definition}
Intuitively, $p_\mdp$ controls the difficulty of observing all internal
transitions, sufficient to recover the strongly connected structure of the true
MECs. Conversely, $B_\mdp$ controls the difficulty of observing MEC-exiting
transitions, eliminating state-action pairs that do not belong to any true MEC. Combining the two arguments yields
$\mecs(\mdp)=\mecs(S,A,\act,\widehat E)$ with high probability.


\begin{restatable}{lemma}{empmecs}
\label{lem:empmecs}
    With probability at least 
    \begin{align*}
        1 - |S|^2|A|\cdot \exp\!\left(-N\cdot p_\mdp\right) - |S||A| \cdot\exp\!\left(-N/B_\mdp\right)
    \end{align*}
    we have $\mecs(\mdp) = \mecs(S,A,\act,\widehat E)$.
\end{restatable}
\lw{Use $\max(B_\mdp, 1)$ to avoid division by zero}
\lw{Can be improved from $|S|^2|A|$ to $|S|$ for the $p_\mdp$ term, but this affects sample complexity only logarithmically.}

\paragraph{Constructing Samples for the Collapsed MDP.}
Assume now that the empirical support correctly identifies
$\mecs(\mdp)$.
It remains to simulate a generative model for $\coll\mdp=\collmdp{\mdp}{F}{F'}$ using the samples from $\mdp$.

The transitions under the distinguished action $\stay$ are deterministic and
therefore require no sampling. For every $\ec\in\mecs(\mdp)$, action
$(s,a)\in\out(\ec)$ and $i\in[N]$, define
\begin{equation}
\label{eq:collsamples}
\coll d_{\ec,(s,a)}^{(i)}
\defeq
\ec_{d_{s,a}^{(i)}}
\end{equation}
where $\ec_{s'}$ denotes the unique MEC containing $s'$.
Crucially, if $d^{(i)}_{s,a}\sim P(\cdot\mid s,a)$, then
$\coll d^{(i)}_{\ec,(s,a)}
\sim
\coll P(\cdot\mid\ec,(s,a))$
by the definition of the collapsed transition kernel (\cref{eq:defcollp}).

\paragraph{Putting the Pieces Together.}
To apply \textsc{C-Reach} on the collapsed MDP, it remains to bound its
transient time $B_{\collmdp\mdp F {F'}}$. By lifting deterministic stationary policies on the
collapsed MDP as in \cref{sec:LP}, we show that the induced transient time increases by at most one, corresponding to the explicit $\stay$-transition into an absorbing state.
\begin{restatable}{lemma}{transientcoll}
\label{lem:transientcoll}
Let $\mdp=(S,A,\act,s_0,P)$ be an MDP and $F,F'\subseteq S$. Then    $B_{\collmdp\mdp F {F'}}\leq B_{\mdp}+1$.
\end{restatable}
Finally, we exploit that \cref{alg:rlreach} returns mixtures of deterministic
stationary policies, lifting each component of the mixture as described in the
preceding section.

We now arrive at the main sample complexity result of this paper.
Combining \cref{lem:empmecs,lem:transientcoll} with the sample complexity
guarantee of \cref{alg:rlreach} (\cref{prop:reachrl}) yields:
\begin{restatable}[Sample Complexity]{theorem}{recsample}
\label{thm:recsample}
    Let $\mdp$ be an MDP $(S,A,\act,s_0,P)$ satisfying $B_\mdp\leq B$ and $p_\mdp\geq p$, and let $F,F'\subseteq S$, $\thresh\in[0,1]$ and $\epsilon\in(0,1)$. If \pref{crec} is feasible then with probability at least
\begin{align*}
    1 - c\cdot|S||A|\cdot\exp\!\left(-c'\cdot{N\epsilon^2}/{B}\right)-|S|^2|A|\cdot\exp\!\left(-N\cdot p\right)
\end{align*}
\cref{alg:rlrec}
 returns an $\epsilon$-optimal solution to \pref{crec},
where $c,c'>0$ are universal constants. 
\end{restatable}
Ignoring logarithmic factors, this corresponds to a sample complexity of
$\tilde{\mathcal O}(1/p+B/\epsilon^2)$ per state-action pair.

\dw{discuss why no $\epsilon$ in $p$ term}

\paragraph{When Bounds on $B_\mdp$ and $p_\mdp$ are Unknown.}
\cref{alg:rlrec} requires an upper bound $B\geq B_\mdp$ on the transient
time together with a target accuracy $\epsilon>0$. Its finite-time guarantee
(\cref{thm:recsample}) additionally depends on a lower bound
$p\leq p_\mdp$. Whilst $\epsilon$ is a user-specified approximation parameter,
suitable values of $B$ and $p$ may not be available in practice. 
More fundamentally, finite-sample guarantees necessarily depend on known
bounds for MDP parameters beyond its size, such as $B_\mdp$ and $p_\mdp$
\citep{Alur:2022,Yang2022}.

This
motivates the following wrapper, which repeatedly invokes
\cref{alg:rlrec} on increasingly large prefixes of the available samples using
progressively larger transient-time bounds and smaller target accuracies.
Suppose that the generative model provides an infinite stream of
samples
$\mathcal D=(d^{(i)}_{s,a})^{i\in\nat}_{s\in S,a\in A}$,
and let
$\mathcal D_N
\defeq
(d^{(i)}_{s,a})^{i\in[N]}_{s\in S,a\in A}$.
For every $N\ge1$, invoke \cref{alg:rlrec} on input
$\mathcal D_N$, $B_N\defeq N^{1/4}$ and
$\epsilon_N\defeq N^{-1/4}$, and let $\policy_N$ denote the resulting policy.

Since $B_N\to\infty$, the transient-time bound is eventually valid, whilst
$\epsilon_N\to0$. Moreover, once $B_N\ge B_\mdp$, the failure probabilities
given by \cref{thm:recsample} are summable. Hence, by the Borel--Cantelli
lemma, only finitely many invocations fail almost surely. It follows that both
the constraint violation and the optimality gap converge to zero almost
surely.
\begin{restatable}[Asymptotic Convergence]{corollary}{asconv}
\label{cor:asconv}
Let $\mdp$ be an MDP $(S,A,\act,s_0,P)$, $F,F'\subseteq S$ and $\thresh\in[0,1]$.
Let $(\policy_N)_{N\ge1}$ be defined as above and suppose $\policy\in\pmix(\mdp)$ satisfies
$\precc\mdp\policy{F'}\geq\thresh$.
Then almost surely,
\begin{align*}
\lim_{N\to\infty}
\max\!\left\{
0,\,
\thresh-\precc\mdp{\policy_N}{F'}
\right\}
&=0,\\
\lim_{N\to\infty}
\max\!\left\{
0,\,
\precc\mdp\policy F
-
\precc\mdp{\policy_N}F
\right\}
&=0.
\end{align*}
\end{restatable}

\section{Nearly Matching Lower Bound}
\label{sec:lower}
\begin{figure}
    \begin{center}
    \scalebox{0.6}{
        \begin{tikzpicture}[node distance = 2cm]
            \node[state, initial right] (s0) {$s_0$};
            \node[state, below left of=s0] (u0) {$u_0$};
            \node[state, accepting, left of=u0] (t0) {$t_0$};
            \node[state, accepting, below right of=s0] (t1) {$t_1$};
            \node[state, right of=t1] (u1) {$u_1$};
              \draw[->] 
              (s0) edge [bend right] node [above left]{$a/\frac{1}{2}$} (t0)
                (t0) edge [loop below] node [below]{$a,b/1$} (t0)
                (s0) edge node [above left]{$a/\frac{1}{2}$} (u0)
                (u0) edge [loop below] node [below]{$a,b/1$} (u0)
                (s0) edge node [below left]{$b/1$} (t1)
                (t1) edge [bend right] node [below]{$a,b/1$} (u1)
                (u1) edge [bend right] node [above]{$b/\beta$} (t1)
                (u1) edge [loop right] node [right]{$a/1, b/1-\beta$} (u1);
        \end{tikzpicture}  
        }  
    \end{center}
    \caption{An unconstrained recurrence objective $F = \{t_0, t_1\}$ where $\beta = 0$ must be distinguished from $\beta \in (0,1]$.}
    \label{fig:reclowerbound}
\end{figure}

The previous section established an upper bound of
$\tilde{\mathcal O}(1/p+B/\epsilon^2)$ samples per state-action pair for \pref{crec}. We now show that a dependence on a parameter beyond $B$ is unavoidable. Indeed, unlike constrained reachability, $B$ is insufficient even for \emph{unconstrained} recurrence.

\cref{fig:reclowerbound} illustrates a fundamental difficulty imposed by MEC identification. Every policy
reaches recurrent behaviour within at most two steps, so transient time is uniformly bounded by $B=2$ irrespective of $\beta \in [0,1]$. If $\beta = 0$, then the only MEC intersecting the target region is $\{t_0\}$. In this case, the optimal action at $s_0$  is $a$ with recurrence probability $1/2$. If $\beta>0$, then $\{t_1,u_1\}$ becomes an additional target MEC with $(u_0,b,t_1)$ an internal transition, and repeatedly selecting action $b$ achieves recurrence probability $1$. Consequently, any algorithm returning a $1/4$-optimal policy must distinguish the cases $\beta=0$ and $\beta>0$. This is in general impossible, and requires $\tilde\Theta(1/p)$ samples under an assumed lower bound $p$ on positive values of $\beta$.

Generalising this construction yields a family of unconstrained recurrence
instances with uniformly bounded transient time which establishes the following
lower bound.

\begin{restatable}[Lower Bound for $p$ Dependence]{theorem}{lowerboundp}
\label{thm:lowerboundp}
For every $n,m\geq 1$ and $p\in(0,1]$, there exists a family of
\emph{unconstrained} recurrence problems $(\mdp,F)$ such that
\begin{enumerate}
    \item every $\mdp=(S,A,\act,s_0,P)$ satisfies
    $|S|=3(n+1)$, $|A|=m+1$, $B_\mdp\le3$, and $p_\mdp\ge p$;
    \item any generative-model RL algorithm that returns
    a $1/8$-optimal policy with probability at least $3/4$ on every problem
    in the family has expected sample complexity
    $\Omega(1/p)$ per state-action pair
    on at least one instance.
\end{enumerate}
\end{restatable}

Combining \cref{thm:lowerboundp} with the reachability lower bound of
$\Omega(B/\epsilon^2)$ yields our
main complexity characterisation, showing the near optimality of our sample complexity result \cref{thm:recsample}.
\begin{restatable}[Combined Lower Bound]{corollary}{maincomplexity}
    \label{cor:maincomplexity}
    Any generative-model RL algorithm for \pref{crec} that returns an $\epsilon$-optimal policy with probability at least $1-\delta$ for all feasible instances $(\mdp,F,F',\thresh)$ satisfying $B_\mdp \leq B$ and $p_\mdp \geq p$ has sample complexity $\Omega(1/p + B/\epsilon^2)$ per state-action pair.
\end{restatable}

\dw{also shows that reduction to limit average without knowing graph cannot be done cheaply}

\section{Concluding Remarks}
\label{sec:conc}
We identified sufficient policy classes for constrained recurrence and established the first near-optimal generative-model sample complexity guarantees. Our results reveal structural parameters governing statistical difficulty and highlight that MEC identification poses a fundamental challenge absent from the average-reward setting. 

\paragraph{Limitations.}
Extensions to the $\omega$-regular setting, multiple constraints, and exact feasibility under assumed strict-feasibility margins are important directions for future work.

\section*{Acknowledgments}
This research is supported by the National Research Foundation, Singapore, under its RSS
Scheme (NRFRSS2022-009).

\bibliography{lit}

\clearpage


\newpage

\appendix
\onecolumn
\dw{check single column is allowed}

\section{Supplementary Materials for \cref{sec:back}}
\label{app:back}

\begin{definition}[Bias Function]
\label{def:bs}
    The \emph{bias} function of a stationary policy $\policy\in\ps(\mdp)$ is
\begin{align*}
    h^{\mdp,\policy}_R(s)\defeq \clim_{T\to\infty}\E_{\run\sim\distrthree\policy{\mdp[s]}}\left[\sum_{t=0}^{T-1} (R(\runstate_t,\runaction_t)-\exavg{\mdp[\runstate_t]} \pi R)\right]
\end{align*}
where $\clim$ denotes the Cesaro-limit.
\end{definition}
\begin{definition}[Bias Span]
    The \emph{bias span} $H_{\mdp,R}$ of $\mdp$ w.r.t.\ $R$ is defined by
\begin{align*}
    H_{\mdp,R}\defeq\max_{\policy\in\ps(\mdp)}\left(\max_{s\in S} h_{R}^{\mdp,\policy}(s)-\min_{s'\in S} h_{R}^{\mdp,\policy}(s')\right)
\end{align*}
\end{definition}

\section{Supplementary Materials for \cref{sec:setup}}
\label{app:setup}

\dw{make sure this is numbered Definition 18!!!}

\begin{definition}[Augmented MDP]
\label{def:mdpaug}
    Let $\mdp=(S,A,\act,s_0,P)$ be an MDP and $F,F'\subseteq S$. The \emph{augmented MDP} $\mdpaug$  is defined by $(S\times\{0,1\},A,\act',(s_0,0),P')$ where $\act'(s,b)\defeq\act(s)$ and
    \begin{align*}
        P'((s',b')\mid (s,b),a)&\defeq
        \begin{cases}
        P(s'\mid s,a)&\tif s'\in F\text{ and } b'=0\\ 
        P(s'\mid s,a)&\tif s'\in F' \setminus F\text{ and } b'=1\\ 
        P(s'\mid s,a)&\tif s'\not\in F\cup F'\text{ and } b'=b\\ 
        0&\tow
    \end{cases}
    \end{align*}
\end{definition}

\section{Supplementary Materials for \cref{sec:reach}}

\spantrans*
\begin{proof}
Let $\policy\in\ps(\mdp)$. Note that for every 
$s\in S$ and $a\in A$, $0\leq R(s,a)\leq\exavg{\mdp[s]}\policy{R_F}\leq 1$
Besides, for every 
state 
$s\in R_\pi$ which is recurrent under $\policy$, $R(s,a)=\ind{s\in A}=\exavg{\mdp[s]}\policy{R_F}$ for all $a\in A$. Hence, for all $s\in S$, 
\begin{align*}
    0\geq h_{R_F}^{\mdp,\policy}(s)\geq\clim_{T\to\infty}\E_{\run\sim\distrthree{\policy}{\mdp[s]}}\left[\sum_{t=0}^{T-1}-\ind{\runstate_t\not\in R_\pi}\right]\geq -B_\mdp
\end{align*}
The claim follows since $\policy\in\ps(\mdp)$ was arbitrary.
\end{proof}

\section{Absorbing MDPs}
\label{app:absorb}

In this appendix we give a self-contained account of absorbing MDPs and we establish some facts which will be used to prove results in \cref{sec:LP}.

\begin{definition}
    Let $\mdp=(S,A,\act,s_0,P)$ be an MDP. A state $s\in S$ is \emph{absorbing} if $P(s\mid a,s)=1$ for all $a\in A$, and $\mdp$ is \emph{absorbing} if absorbing states are reachable from any state under any stationary policy, i.e.\  for all $s\in S$ and $\pi\in\ps(\mdp)$,
    \begin{align*}
        \prob_{\run\sim\distrthree\policy{\mdpsub s}}\left[\exists i\in\nat\ldotp\runstate_i\in T\right]>0
    \end{align*}
    where $T\subseteq S$ is the set of absorbing states.
\end{definition}

The following is immediate from the definition of absorbing MDP (see the proof of \cref{lem:abs2} for a similar argument).
 \begin{lemma}
    \label{lem:uniformbound}
    Let $\mdp$ be an absorbing MDP, $\policy\in\pmix(\mdp)$, $s \in S \setminus T$ and $a \in \act(s)$, we have
    \begin{align*}
        \E_{\run \sim \distr} \left[ \sum_{i=0}^\infty \ind{ \runstate_i = s, \runaction_i = a} \right] < \infty
    \end{align*}
\end{lemma}

\subsection{Flow Equations for Absorbing MDPs}
\label{app:lp}
For absorbing $\mdp=(S,A,\act,s_0,P)$, expected state-action occupation counts (aka occupancy measures) $x(s,a)$ can be axiomatised via the following flow equations:
\begin{align}
&
\tag{Flow}
\label{eq:pres}
\sum_{a \in\act(s)} x(s,a)=
\ind{s=s_0}+\sum_{s' \in S\setminus T}
\sum_{a \in\act(s')}
P(s \mid s',a) \cdot x(s',a)
&
(s \in S\setminus T)\\
&
\notag
x(s,a)\ge0 & (s\in S\setminus T, a\in\act(s))
\end{align}
Non-negative solutions of \eqref{eq:pres} give rise to stationary stochastic policies $\policy(a\mid s)\propto x(s,a)$:
\begin{lemma}
    \label{lem:lptopolicy}
    Let $\mdp$ be an absorbing MDP. If $x: (S \setminus T) \times A \to \real_{\geq 0}$ satisfies \eqref{eq:pres}, then there exists $\policy\in\ps(\mdp)$ satisfying
    \begin{align*}
        \E_{\run \sim \distr} \left[ \sum_{i \in \nat} \ind{ \runstate_i = s, \runaction_i = a} \right] &= x(s,a)
    \end{align*}
    for all $s \in S \setminus T$ and $a \in \act(s)$,
    and for all $s\in S \setminus T$,
    \begin{align*}
        \sum_{a \in \act(s)} \ind{ \policy(a \mid s) > 0} &= \max \left\{ 1, \sum_{a \in A} \ind{ x(s,a) > 0 } \right\}
    \end{align*}
\end{lemma}
\begin{proof}
    Let $x$ be a non-negative solution to \eqref{eq:pres}. For each $s \in S$, fix an arbitrary $a_s \in \act(s)$ and define
    \begin{align*}
        \policy(a \mid s) &\defeq \begin{cases}
            \frac{x(s,a)}{\sum_{a' \in \act(s)} x(s,a')} & s \notin T \text{ and } \sum_{a' \in \act(s)} x(s,a') > 0 \\
            \ind{a = a_s} & \text{otherwise}
        \end{cases} && s \in S, a \in \act(s)
    \end{align*}
    Let $s \in S \setminus T$ and $a \in \act(s)$. First, suppose $\sum_{a' \in \act(s)} x(s,a') > 0$. Using \eqref{eq:pres}, we observe
    \begin{align*}
        &\policy(a \mid s) \cdot \left( \ind{s=s_0} + \sum_{s' \in S \setminus T} \sum_{a' \in \act(s')} P(s \mid s',a') \cdot x(s',a') \right) \\
        &= \frac{x(s,a)}{\sum_{a' \in \act(s)} x(s,a')} \cdot \left( \ind{s=s_0} + \sum_{s' \in S \setminus T} \sum_{a' \in \act(s')} P(s \mid s',a') \cdot x(s',a') \right) \\
        &= x(s,a) \cdot \frac{ \ind{s=s_0} + \sum_{s' \in S \setminus T} \sum_{a' \in \act(s')} P(s \mid s',a') \cdot x(s',a') }{\sum_{a' \in \act(s)} x(s,a')} \\
        &= x(s,a) \cdot \frac{ \ind{s=s_0} + \sum_{s' \in S \setminus T} \sum_{a' \in \act(s')} P(s \mid s',a') \cdot x(s',a') }{ \ind{s=s_0} + \sum_{s' \in S \setminus T} \sum_{a' \in \act(s')} P(s \mid s',a') \cdot x(s',a') } \\
        &= x(s,a)
    \end{align*}
    Now suppose otherwise that $\sum_{a' \in \act(s)} x(s,a') = 0$. Then
    \begin{align*}
        x(s,a)= 0 = \sum_{a' \in \act(s)} x(s,a') = \ind{s=s_0} + \sum_{s' \in S \setminus T} \sum_{a' \in \act(s')} P(s \mid s',a') \cdot x(s',a'),
    \end{align*}
    where we again use \eqref{eq:pres}. Thus, in either case, we have
    \begin{align*}
        x(s,a) = \policy(a \mid s) \cdot \left( \ind{s=s_0} + \sum_{s' \in S \setminus T} \sum_{a' \in \act(s')} P(s \mid s',a') \cdot x(s',a') \right),
    \end{align*}
    Hence, \cref{cor:actoccupancy} below yields
    \begin{align*}
        x(s,a) = \E_{\run \sim \distr} \left[ \sum_{i \in \nat} \ind{ \runstate_i = s, \runaction_i = a } \right]
    \end{align*}
    for all $s \in S \setminus T$ and $a \in \act(s)$
\end{proof}

\begin{lemma}
    \label{lem:policytolp}
    Let $\policy \in \pmix(\mdp)$. Then the following assignment satisfies \cref{eq:pres}:
    \begin{align*}
        x(s,a)\defeq &\E_{\run \sim \distr} \left[ \sum_{i=0}^\infty \ind{\runstate_i = s, \runaction_i = a} \right] && s \in S \setminus T, a \in\act(s)
    \end{align*}    
\end{lemma}
\begin{proof}
    Fix $s \in S \setminus T$. We verify
    \begin{align*}
        \sum_{a \in \act(s)} x(s,a) &= \sum_{a \in \act(s)} \E_{\run \sim \distr} \left[ \sum_{i=0}^\infty \ind{\runstate_i = s, \runaction_i = a} \right] = \E_{\run \sim \distr} \left[ \sum_{i=0}^\infty \ind{\runstate_i = s} \right] \\
        &= \ind{s=s_0} + \sum_{i \in \nat} \prob_{\run \sim \distr} \left[ \runstate_{i+1} = s \right] \\
        &= \ind{s=s_0} + \sum_{i \in \nat} \sum_{s' \in S \setminus T} \sum_{a \in \act(s')} \prob_{\run \sim \distr} \left[ \runstate_i = s', \runaction_i= a, \runstate_{i+1} = s \right] \\
        &= \ind{s=s_0} + \sum_{i \in \nat} \sum_{s' \in S \setminus T} \sum_{a \in \act(s')} P(s \mid s',a) \cdot \prob_{\run \sim \distr} \left[ \runstate_i = s', \runaction_i= a \right] \\
        &= \ind{s=s_0} + \sum_{s' \in S \setminus T} \sum_{a \in \act(s')} P(s \mid s',a) \cdot \sum_{i \in \nat} \prob_{\run \sim \distr} \left[ \runstate_i = s', \runaction_i= a \right] \\
        &= \ind{s=s_0} + \sum_{s' \in S \setminus T} \sum_{a \in \act(s')} P(s \mid s',a) \cdot \E_{\run \sim \distr} \left[ \sum_{i=0}^\infty \ind{\runstate_i = s', \runaction_i = a} \right] \\
        &= \ind{s=s_0} + \sum_{s' \in S \setminus T} \sum_{a \in \act(s')} P(s \mid s',a,) \cdot x(s',a)
    \end{align*}
    as needed.
\end{proof}

\begin{lemma}[{\cite[Thm.~11.4]{grinstead1997introduction}}]
    \label{lem:stateoccupancy}
    Let $\mc = (S, s_0, P)$ be an absorbing Markov chain (i.e. an absorbing MDP with $|A| = 1$). Then the map
    \begin{align*}
        s \mapsto \E_{\run \sim \distrtwo{}} \left[ \sum_{i \in \nat} \ind{ \runstate_i = s} \right] && s \in S \setminus T
    \end{align*} is the unique $x: S \setminus T \to \real$ satisfying
    \begin{align*}
        x(s) = \ind{s=s_0} + \sum_{s' \in S \setminus T} P(s \mid s') \cdot x(s') && s \in S \setminus T.
    \end{align*}
\end{lemma}

\begin{corollary}
    \label{cor:actoccupancy}
    Let $\mc = (S,A,\act,s_0,P)$ be an absorbing MDP and $\policy$ be stationary. Then
    \begin{align*}
        x(s,a) &\defeq \E_{\run \sim \distr} \left[ \sum_{i \in \nat} \ind{ \runstate_i = s, \runaction_i = a} \right] && s \in S \setminus T, a \in \act(s)
    \end{align*} is the unique $x: \{(s,a)\in (S \setminus T) \times A\mid a\in\act(s)\} \to \real$ satisfying
    \begin{align*}
        x(s,a) &= \policy(a \mid s) \cdot \left( \ind{s=s_0} + \sum_{s' \in S \setminus T} \sum_{a' \in \act(s')} P(s \mid s',a') \cdot x(s',a') \right) && s \in S \setminus T, a \in \act(s).
    \end{align*}
\end{corollary}
\begin{proof}
    The "is a solution" direction follows from \cref{lem:policytolp}. Define the Markov chain $\mc_\policy \defeq (S,s_0,P_\policy)$ with $P_\policy$ given by
    \begin{align*}
        P_\policy(s' \mid s) &\defeq \sum_{a \in \act(s)} \policy(a \mid s) \cdot P(s' \mid s,a) && s,s' \in S.
    \end{align*}
    Let $T_\policy$ denote the absorbing states of $\mc_\policy$:
    \begin{align*}
        T_\policy \defeq \{ s \in S \mid P_\policy(s \mid s) = 1 \} &= \left\{ s \in S \mid \sum_{a \in \act(s)} \policy(a \mid s) \cdot P(s \mid s,a) = 1  \right\}.
    \end{align*}
    It is evident that $T \subseteq T_\policy$. Furthermore, if there was $s \in T_\policy \setminus T$, we would have
    \begin{align*}
        \prob_{\run \sim \distrthree{\policy}{\mdpsub s}} \left[ \exists i \in \nat. \runstate_i \in T \right] = 0,
    \end{align*}
    which contradicts $\mdp$ being absorbing.  Thus we must instead have $T_\policy = T$. By construction, we have
    \begin{align*}
        \prob_{\run \sim \distrthree{}{\mc_\policy[s]}} \left[ \exists i \in \nat. \runstate_i \in T \right] &= \prob_{\run \sim \distrthree{\policy}{\mdpsub s}} \left[ \exists i \in \nat. \runstate_i \in T \right] > 0 
    \end{align*}
    for all $s\in S$ since $\mdp$ is absorbing. Thus $\mc_\policy$ is also absorbing.
    Now suppose $x: (S \setminus T) \times A \to \real$ satisfies
    \begin{align}
        x(s,a) &= \policy(a \mid s) \cdot \left( \ind{s=s_0} + \sum_{s' \in S \setminus T} \sum_{a' \in \act(s')} P(s \mid s',a') \cdot x(s',a') \right) \label{eq:flowsat}
    \end{align}
    for all $s \in S \setminus T, a \in \act(s)$.
    We define
    \begin{align*}
        \overline{x}(s) &\defeq \sum_{a' \in \act(s)} x(s,a') && s \in S \setminus T.
    \end{align*}
    Then for $s \in S \setminus T$, using \cref{eq:flowsat}, we have
    \begin{align}
    \notag
        \overline{x}(s) = \sum_{a' \in \act(s)} x(s,a') &= \sum_{a' \in \act(s)} \policy(a' \mid s) \cdot \left( \ind{s=s_0} + \sum_{s' \in S \setminus T} \sum_{a \in \act(s')} P(s \mid s',a) \cdot x(s',a) \right) \\
        &= \ind{s=s_0} + \sum_{s' \in S \setminus T} \sum_{a \in \act(s')} P(s \mid s',a) \cdot x(s',a) \label{eq:flowsat2}
    \end{align}
    Combining \cref{eq:flowsat} and \cref{eq:flowsat2},
    \begin{align}
        x(s,a) &= \policy(a \mid s) \cdot \overline{x}(s) \label{eq:flowsat3}
    \end{align}
    for all $s \in S \setminus T, a \in \act(s)$.
    Combining \cref{eq:flowsat2,eq:flowsat3} and the definition of $P_\policy$,
    \begin{align*}
        \overline{x}(s) &= \ind{s=s_0} + \sum_{s' \in S \setminus T} \sum_{a \in \act(s')} P(s \mid s',a) \cdot \policy(a \mid s') \cdot \overline{x}(s') \\
        &= \ind{s=s_0} + \sum_{s' \in S \setminus T} \left( \sum_{a \in \act(s')} \policy(a \mid s') \cdot P(s \mid s', a) \right) \cdot \overline{x}(s') \\
        &= \ind{s=s_0} + \sum_{s' \in S \setminus T} P_\policy(s \mid s') \cdot \overline{x}(s')
    \end{align*}
    for $s \in S \setminus T$. By uniqueness (\cref{lem:stateoccupancy}), we must have
    \begin{align*}
        \overline{x}(s) = \E_{\run \sim \distr} \left[ \sum_{i \in \nat} \ind{ \runstate_i = s} \right]
    \end{align*}
    for all $s \in S \setminus T$.
    Thus, we conclude
    \begin{align*}
        x(s,a) &= \policy(a \mid s) \cdot \overline{x}(s) \\
        &= \policy(a \mid s) \cdot \E_{\run \sim \distr} \left[ \sum_{i \in \nat} \ind{ \runstate_i = s} \right] \\
        &= \E_{\run \sim \distr} \left[ \sum_{i \in \nat} \ind{ \runstate_i = s, \runaction_i = a} \right]
    \end{align*}
    for $s \in S \setminus T$ and $a \in \act(s)$ as needed.
\end{proof}

\section{Supplementary Materials for \cref{sec:LP}}

We recall the following well-known fact about end components:
\begin{lemma}[{\citep[Thm.~3.2]{Alfaro:99}}]
    \label{lem:alfaro}
    Let $\mdp = (S,A,\act,s_0,P)$ be an MDP and $\policy \in \pmix(\mdp)$.
    For a run $\run$ define
    \begin{align*}
        S_{\ecr\run}&\defeq\infi(\run)&
        \act_{\ecr\run}(s)\defeq&\{a\in A\mid\text{for infinitely many $i\in\nat$. $\runstate_i=s$ and $\runaction_i=a$}\}
    \end{align*}
    and $\ecr\run\defeq(S_{\ecr\run},\act_{\ecr\run})$.
    Then
    \begin{align*}
        \prob_{\run\sim\distr}[\ecr\run\text{ is an end component}]=1
    \end{align*}
\end{lemma}

\subsection{Collapsed MDP}

\begin{definition}
Let $\mdp=(S,A,\act,s_0,P)$ be an MDP, let $F,F'\subseteq S$, and let
$\coll s_0$ denote the unique MEC containing $s_0$. The
\emph{collapsed MDP} $\collmdp{\mdp}{F}{F'}$ is the MDP
$(\coll S,\coll A,\coll\act,\coll s_0,\coll P)$, where
\begin{align*}
\coll S
    &\defeq
    \mecs(\mdp)\cup
    \{\bot,\bot_F,\bot_{F'},\bot_{F\land F'}\},\\
\coll A
    &\defeq
    \bigcup_{\ec\in\mecs(\mdp)}\out(\ec)
    \cup\{\stay\},\\
\coll\act(\ec)
    &\defeq
    \begin{cases}
        \out(\ec)\cup\{\stay\}
            & \tif\ec\in\pmecs(\mdp)\\
        \out(\ec)
            & \tif\ec\in\mecs(\mdp)\setminus\pmecs(\mdp)\\
        \{\stay\}
            & \tow.
    \end{cases}
\end{align*}
The transition kernel $\coll P$ is defined as follows. For every
$\ec,\ec'\in\mecs(\mdp)$ and $(s,a)\in\out(\ec)$,
\begin{align*}
    \coll P(\ec'\mid\ec,(s,a))
\defeq
\sum_{s'\in S_{\ec'}}P(s'\mid s,a).
\end{align*}
Moreover, every absorbing state in
$\{\bot,\bot_F,\bot_{F'},\bot_{F\land F'}\}$ has a deterministic self-loop under
$\stay$. Finally, for every proper MEC $\ec\in\pmecs(\mdp)$, executing $\stay$
causes a deterministic transition to the absorbing state $\bot_\ec$, where
\begin{align}
\label{eq:botec}
\bot_\ec\defeq
\begin{cases}
\bot_{F\land F'}
&
\tif S_\ec\cap F\neq\emptyset
\text{ and }S_\ec\cap F'\neq\emptyset\\
\bot_F
&
\tif S_\ec\cap F\neq\emptyset
\text{ and }S_\ec\cap F'=\emptyset\\
\bot_{F'}
&
\tif S_\ec\cap F=\emptyset
\text{ and }S_\ec\cap F'\neq\emptyset\\
\bot
&
\tow
\end{cases}
\end{align}    
\end{definition}

\begin{lemma}
\label{lem:abs}
 Let $\mdp=(S,A,\act,s_0,P)$ be an MDP and $F,F'\subseteq S$. Then $\collmdp\mdp F{F'}$ is absorbing.
\end{lemma}
\begin{proof}
Suppose $\coll\mdp=(\coll S,\coll A,\coll\act,\coll s_0,\coll P)=\collmdp\mdp F{F'}$.
    Observe that $\bot,\bot_F,\bot_{F'}$ and $\bot_{F\land F'}$ are the only absorbing state in $\coll\mdp$ and assume towards contradiction that there exists $s\in \coll S$ and $\policy\in\ps(\coll\mdp)$ satisfying
    \begin{align*}
        \prob_{\run\sim\distrthree\policy{\coll\mdp[s]}}\left[\exists i\in\nat\ldotp \runstate_i\in\{\bot,\bot_F,\bot_{F'},\bot_{F\land F'}\}\right]=0
    \end{align*}
    Then there exists an EC $\coll\ec$ for $\coll\mdp$ satisfying $\{\bot,\bot_F,\bot_{F'},\bot_{F\land F'}\}\cap S_{\coll\ec}=\emptyset$. Note that by definition of $\coll\mdp$ and end components, $S_{\coll\ec}$ cannot be a singleton.  Now consider,
    \begin{align*}
        S_{\ec'} &\defeq \bigcup_{\ec''\in S_{\coll\ec}} S_{\ec''}\subseteq S&
        \act_{\ec'}(s)\defeq \act_{\ec_s}(s)\cup\{a\in A\mid (s,a)\in\act_{\coll\ec}(\ec_s)\}\subseteq \act(s)
    \end{align*}
    where $s\in S_{\ec'}\subseteq S$  and $\ec_s\in S_{\coll\ec}$ is the unique MEC in $\mdp$ containing.

    By construction $\ec'\defeq (S_{\ec'}, \act_{\ec'})$ is an EC, and it strictly extends any EC in $S_{\coll\ec}$, which is a contradiction to their supposed maximality.
\end{proof}

\subsection{Policy Lifting}

\begin{definition}[Deterministic Lift]
\label{def:liftdet}
    For every proper MEC $\ec=(S_\ec,\act_\ec)$, every non-empty set
$T\subseteq S_\ec$, and every state $s\in S_\ec$, fix an action
$a_{s,T}\in\act_\ec(s)$ that is the first action on a shortest path in the
graph $(S_\ec,\rightarrow_{\act_\ec})$ from $s$ to some state in $T$.

    For $\policy\in\pd(\mdp)$ we define its \emph{deterministic lift} $\liftdet\policy\in\pd(\mdpaug)$ as follows:
    Let $(s,b)$ be a state of $\mdpaug$, where $b\in\{0,1\}$, and let
$\ec=(S_\ec,\act_\ec)$ be the unique MEC containing $s$. If $\policy(\ec)=(s',a)\in\out(\ec)$, then
\begin{align*}
    \liftdet\policy(s,b)\defeq
\begin{cases}
a_{s,\{s'\}} & \tif s\neq s'\\
a & \tow
\end{cases}
\end{align*}
Otherwise, $\policy(\ec)=\stay$, and
\begin{align*}
    \liftdet\policy(s,b)\defeq
\begin{cases}
a_{s,S_\ec\cap F}&\tif S_\ec\cap F\neq\emptyset,\;S_\ec\cap F'=\emptyset,\\
a_{s,S_\ec\cap F'}&\tif S_\ec\cap F=\emptyset,\;S_\ec\cap F'\neq\emptyset,\\
a_{s,S_\ec\cap F\cap F'}&\tif S_\ec\cap F\cap F'\neq\emptyset,\\
a_{s,S_\ec\cap F'}&\tif S_\ec\cap F\neq\emptyset,\; S_\ec\cap F'\neq\emptyset,\;S_\ec\cap F\cap F'=\emptyset,\; b=0\\
a_{s,S_\ec\cap F}&\tif S_\ec\cap F\neq\emptyset,\; S_\ec\cap F'\neq\emptyset,\; S_\ec\cap F\cap F'=\emptyset,\; b=1
\end{cases}
\end{align*}
\end{definition}
Intuitively, $\liftdet\policy$ follows the exit decisions prescribed by
$\policy$. If $\policy$ commits to remaining within a MEC by selecting
$\stay$, then $\liftdet\policy$ repeatedly steers the process towards the
relevant target set inside the MEC. When the MEC intersects both $F$ and
$F'$ but not their intersection, the augmentation bit records which target set
was visited most recently and is used to alternate between $F$ and $F'$.
Consequently, both sets are visited infinitely often almost surely.

\absone*
\begin{proof}
    Let $\policy\in\pd(\coll\mdp)$ and let $\lift\policy\in\ps(\mdp)$ be its lift.
For a run $\run\sim\distrthree{\lift\policy}\mdp$ define inductively the embedded process $(\run^{\coll S}_i)_{i\ge0}$ in $\coll S$
together with the corresponding transition times $(\tau_i)_{i\ge0}$:
\begin{align*}
    \tau_0&\defeq 0&
\run^{\coll S}_0&\defeq\coll s_0
\end{align*}
For $i\geq 1$, if $\run^{\coll S}_{i-1}\in\{\bot,\bot_F,\bot_{F'},\bot_{F\land F'}\}$ then
\begin{align*}
    \tau_i&\defeq\tau_{i-1}&
    \run^{\coll S}_i&\defeq\run^{\coll S}_{i-1}
\end{align*}
Otherwise, $\run^{\coll S}_{i-1}\in\mecs(\mdp)$. If $\policy(\run^{\coll S}_{i-1})=\stay$ then we define
\begin{align*}
    \tau_i&\defeq\tau_{i-1}&
    \run^{\coll S}_i&\defeq\bot_{\run^{\coll S}_{i-1}}
\end{align*}
(cf.~\cref{eq:botec}). Otherwise, $\policy(\run^{\coll S}_{i-1})=(s,a)\in\out(\run^{\coll S}_{i-1})$ and we define
\begin{align*}
    \tau_{i}
&\defeq
1+\inf\{t\geq\tau_{i-1}\mid
\runstate_t=s,
\runaction_t=a
\}
\end{align*}
and $\run^{\coll S}_{i}$ is the unique MEC containing $\runstate_{\tau_i}$.
Note that $\tau_i<\infty$ almost surely. Indeed, until the first visit to $s$, $\lift\policy$ coincides with the uniform random policy on the MEC $\run^{\coll S}_{i-1}$. Since the induced Markov chain is finite and irreducible, every state is reached almost surely.
Likewise,
\begin{align}
\label{eq:ECinf}
    \prob_{\run\sim\distrthree{\lift\policy}\mdp}\left[\infi(\run)\cap F\neq\emptyset\right]=\prob_{\run\sim\distrthree{\lift\policy}\mdp}\left[\exists i\geq 0\ldotp \run^{\coll S}_i\in\{\bot_F,\bot_{F,F'}\}\right]
\end{align}
\dw{not completely obvious}
By definition of the embedded process and \cref{eq:defcollp},
\begin{align}
\label{eq:markov}
    \prob_{\run\sim\distrthree{\lift\policy}\mdp}\left[\run^{\coll S}_{i+1}=\coll s'\mid\run^{\coll S}_{i}=\coll s_i\right]=\coll P\left(\coll s'\mid \coll s,\policy(\coll s)\right)
\end{align}
for all $\coll s,\coll s'\in\coll S$ and $i\geq 0$ satisfying
 $\prob_{\run\sim\distrthree{\lift\policy}\mdp}[\run^{\coll S}_{i}=\coll s]>0$.

Hence, using \cref{eq:ECinf,eq:markov},
 \begin{align*}
     \precc\mdp{\lift\policy} F&=\prob_{\run\sim\distrthree{\lift\policy}\mdp}\left[\infi(\run)\cap F\neq\emptyset\right]\\
    &=\prob_{\run\sim\distrthree{\lift\policy}\mdp}\left[\exists i\geq 0\ldotp \run^{\coll S}_i\in\{\bot_F,\bot_{F,F'}\}\right]\\
    &=\prob_{\coll\run\sim\distrthree{\policy}{\coll\mdp}}\left[\exists i\geq 0\ldotp \coll\run^{\coll S}_i\in\{\bot_F,\bot_{F,F'}\}\right]\\
    &=\precc{\coll\mdp}{\policy}{\{\bot_F,\bot_{F,F'}\}}
 \end{align*}
 The proof for the deterministic lift $\liftdet\policy\in\pd(\mdpaug)$, see \cref{def:liftdet}, is analogous. For the embedded process the augmentation bit is ignored and the selection of actions on the shortest path to leaving actions $(s,a)$ ensures that $\tau_i$ is likewise finite. Moreover, using the augmentation bit, we have
 \begin{align*}
     \prob_{\run\sim\distrthree{\liftdet\policy}{\mdpaug}}\left[\infi(\run)\cap \left(F\times\{0,1\}\right)\neq\emptyset\mid \exists i\geq 0\ldotp \run^{\coll S}_i\in\{\bot_F,\bot_{F,F'}\}\right]=1
 \end{align*}
 as an analogue to \cref{eq:ECinf}.
\end{proof}

\subsection{Projecting Policies to the Collapsed MDP}

\proj*
The proof relies on the following auxiliary lemma (proven below) and \dw{theory of absorbing MDP}:
\begin{lemma}
\label{lem:abs2}
    Let $\policy\in\pmix(\mdp)$. Define
    \begin{align*}
        x(\ec,(s,a))&\defeq\mathbb{E}_{\run \sim \distr}\left[\sum_{i\in\nat}\ind{\runstate_i=s,\runaction_i=a}\right]<\infty&(\ec\in\mecs(\mdp),(s,a)\in \out(\ec)) \\
        x(\ec,\stay)&\defeq\prob_{\run \sim \distr}\left[\exists i\in\nat\ldotp\forall j\geq i\ldotp\runstate_j\in S_\ec\right]        &(\ec\in\pmecs(\mdp))
    \end{align*}
    Then $x$ satisfies \eqref{eq:pres}.
\end{lemma}

\begin{proof}[Proof of \cref{lem:proj}]
Let $\policy\in\pmix(\mdp)$. By \cref{lem:abs2,lem:lptopolicy} there exists $\coll\policy\in\ps(\coll\mdp)$ satisfying
\begin{align*}
    \E_{\run \sim \distrthree{\coll\policy}{\coll\mdp}} \left[ \sum_{i \in \nat} \ind{ \runstate_i = \ec, \runaction_i = \stay} \right]=\prob_{\run \sim \distr}\left[\exists i\in\nat\ldotp\forall j\geq i\ldotp\runstate_j\in S_\ec\right]
\end{align*}
for all $\ec\in\pmecs(\mdp)$. 
Let $\mecs_F(\mdp)\subseteq\pmecs(\mdp)$ be the set of proper MECs intersecting $F$.
By definition of $\coll\mdp$,
\begin{align*}
    \preach{\coll\mdp}{\coll\policy}{\{\bot_F,\bot_{F\land F'}\}}=\sum_{\ec\in\mecs_F(\mdp)}\E_{\run \sim \distrthree{\coll\policy}{\coll\mdp}} \left[ \sum_{i \in \nat} \ind{ \runstate_i = \ec, \runaction_i = \stay} \right]
\end{align*}
Moreover, as a consequence of \cref{lem:alfaro},
\begin{align*}
    \precc\mdp\policy F\leq\sum_{\ec\in\mecs_F(\mdp)}\prob_{\run \sim \distr}\left[\exists i\in\nat\ldotp\forall j\geq i\ldotp\runstate_j\in S_\ec\right]
\end{align*}
which proves the first inequality. The second follows analogously.
\end{proof}

\begin{proof}[Proof of \cref{lem:abs2}]
First, to verify that
$x(\ec,(s,a))<\infty$ for all $\ec\in\mecs(\mdp)$ and $(s,a)\in \out(\ec)$, we
define the abbreviation $\mu\defeq P(\cdot\mid s,a)$ and let
    \begin{align*}
r\defeq\sup_{\policy\in\pmix(\mdp)}\prob_{\run\sim\distrthree\policy{\mdp[\mu]}}\left[\exists i\geq 0\ldotp \runstate_i\in S_\ec\right]
    \end{align*}
    be the optimal probability to return to $\ec$ from initial state distribution $\mu$ \changed[dw]{(slightly abusing notation)}. Note that since this is a reachability problem, the supremum is attained by a deterministic stationary policy and since $\ec$ is maximal, $r<1$ (by analogous reasoning  as in \cref{lem:abs}). Hence,
    \begin{align*}
        x(\ec,(s,a))=\mathbb{E}_{\run \sim \distr}\left[\sum_{i\in\nat}\ind{\runstate_i=s,\runaction_i=a}\right]\leq\sum_{i\in\nat} r^{i}=\frac 1{1-r}<\infty
    \end{align*}

Next, since policies almost surely eventually remain within \changed[dw]{one proper EC, hence}  one proper MEC (equivalently, $\sum_{\ec \in \pmecs(\mdp)} x(\ec,\stay) = 1)$, we have for $\ec\in\mecs(\mdp)$,
    \begin{align}
    \label{eq:assec}
        \lim_{i \to \infty} \prob_{\run \sim \distr} \left[ \runstate_i \in S_\ec \right] &= \begin{cases}
            x(\ec,\stay) & \tif \ec \in \pmecs(\mdp)\\
            0&\tow
        \end{cases}
    \end{align}
    Next, observe that for all runs $\run$, $\ec \in \mecs(\mdp)$, and $i \in \nat$,
    \begin{align*}
        \ind{\runstate_i \in S_\ec} + \sum_{j < i} \ind{ \runstate_j \in S_\ec, \runstate_{j+1} \notin S_\ec } = \ind{\runstate_0 \in S_\ec} + \sum_{j < i} \ind{\runstate_j \notin S_\ec, \runstate_{j+1} \in S_\ec }.
    \end{align*}
    That is, up to time $i$, there is a correspondence between the number of times each $\ec$ was entered and exited in run $\run$. Hence we have
    \begin{align*}
         \ind{\runstate_i \in S_\ec} + \sum_{j < i} \ind{ (\runstate_j,\runaction_j) \in\out(\ec) } &= \ind{\runstate_i \in S_\ec} + \sum_{j < i} \ind{ (\runstate_j, \runaction_j) \in \out(\ec), \runstate_{j+1} \notin S_\ec } + \sum_{j < i} \ind{(\runstate_j, \runaction_j) \in \out(\ec), \runstate_{j+1} \in S_\ec} \\
         &= \left( \ind{\runstate_i \in S_\ec} + \sum_{j < i} \ind{ \runstate_j \in S_\ec, \runstate_{j+1} \notin S_\ec } \right) + \sum_{j < i} \ind{(\runstate_j, \runaction_j) \in \out(\ec), \runstate_{j+1} \in S_\ec} \\
         &= \left( \ind{\runstate_0 \in S_\ec} + \sum_{j < i} \ind{\runstate_j \notin S_\ec, \runstate_{j+1} \in S_\ec } \right) + \sum_{j < i} \ind{(\runstate_j, \runaction_j) \in \out(\ec), \runstate_{j+1} \in S_\ec} \\
         &= \ind{\runstate_0 \in S_\ec} + \sum_{j < i} \ind{(\runstate_j,\runaction_j)\in\out, \runstate_{j+1} \in S_\ec }
    \end{align*}
    where $\out\defeq\bigcup_{\ec\in\mecs(\mdp)}\out(\ec)$.
    Taking expectation over $\run$ yields 
    \begin{align*}
        \prob_{\run \sim \distr}\left[\runstate_i \in S_\ec\right] + \E_{\run \sim \distr} \left[\sum_{j < i} \ind{(\runstate_j,\runaction_j) \in\out(\ec) }\right] =  \ind{s_0\in S_\ec} + \E_{\run \sim \distr} \left[\sum_{j < i} \ind{(\runstate_j,\runaction_j)\in\out, \runstate_{j+1} \in S_\ec }\right]
    \end{align*}
    We now inspect the limiting behaviour of both sides of this equation as $i \to \infty$. By \cref{eq:assec}, for proper $\ec$ (and similarly for improper $\ec$),
    \begin{align*}
        &\lim_{i \to \infty} \left( \prob_{\run \sim \distr}\left[\runstate_i \in S_\ec\right] + \E_{\run \sim \distr} \left[\sum_{j < i} \ind{ (\runstate_j,\runaction_j) \in\out(\ec) } \right] \right) \\
        &= x(\ec,\stay) + \E_{\run \sim \distr} \left[\sum_{i \in \nat} \ind{ (\runstate_i,\runaction_i) \in\out(\ec) }\right] \\
        &= x(\ec,\stay) + \sum_{(s,a)\in\out(\ec)} \underbrace{\E_{\run \sim \distr} \left[\sum_{i \in \nat} \ind{ (\runstate_i,\runaction_i) =(s,a) }\right]}_{x(\ec,(s,a))} \\
        &= \sum_{\coll a\in\coll\act(\ec)} x(\ec,\coll a)
    \end{align*}
    Similarly, we have 
    \begin{align*}
        &\lim_{i \to \infty} \left( \ind{s_0\in S_\ec} + \E_{\run \sim \distr} \left[\sum_{j < i} \ind{ (\runstate_j,\runaction_j)\in\out, \runstate_{j+1} \in S_\ec }\right] \right) \\
        &= \ind{s_0\in S_\ec} + \E_{\run \sim \distr} \left[\sum_{i \in \nat} \ind{ (\runstate_i,\runaction_i)\in\out, \runstate_{i+1} \in S_\ec }\right] \\
        &= \ind{s_0\in S_\ec} + \sum_{\ec' \in \mecs} \sum_{(s,a) \in\out(\ec') } \E_{\run \sim \distr} \left[\sum_{i \in \nat} \ind{ (\runstate_i,\runaction_i)=(s,a), \runstate_{i+1} \in S_\ec }\right] \\
        &= \ind{s_0\in S_\ec} + \sum_{\ec' \in \mecs} \sum_{(s,a) \in\out(\ec') } \underbrace{\E_{\run \sim \distr} \left[\sum_{i \in \nat} \ind{ (\runstate_i,\runaction_i)=(s,a)}\right]}_{x(\ec',(s,a))}\cdot \coll P(\ec\mid \ec',(s,a)) \\
        &= \ind{\ec=\coll s_0} + \sum_{\ec'\in \coll S\setminus \coll T} \sum_{\coll a\in\coll\act(\ec')} \coll P(\ec\mid \ec',\coll a) \cdot x(\ec',\coll a)
    \end{align*}
    \changed[dw]{since $\coll P(\ec\mid\ec',\coll a)>0$ only if $\coll a\neq\stay$.}
    Thus we conclude
    \begin{align*}
        \sum_{\coll a\in\coll\act(\ec)} x(\ec,\coll a) &= \ind{s_0\in S_\ec} + \sum_{\coll s'\in \coll S\setminus \coll T} \sum_{\coll a\in\coll\act(\coll s')} \coll P(\ec\mid \coll s',\coll a) \cdot x(\coll s',\coll a)
    \end{align*} 
    for all $\ec \in\mecs(\mdp)= \coll S\setminus\{\bot,\bot_F,\bot_{F'},\bot_{F\land F'}\}$ as needed.
\end{proof}

\subsection{Sufficient Policy Classes}

\begingroup
\renewcommand{\theenumi}{\arabic{enumi}}
\renewcommand{\labelenumi}{\theenumi.}
\typepolicy*
\endgroup

\begin{proof}[Proof of \cref{thm:type}]
Let $\mdp=(S,A,\act,s_0,P)$ be an MDP, $F,F'\subseteq S$ and $\thresh\in[0,1]$.
For the first part, suppose $\policy\in\pmix(\mdp)$ satisfies
\begin{align*}
\precc\mdp\policy{F'}\geq\thresh
\end{align*}
By \cref{lem:abs2,lem:lptopolicy} \dw{elaborate} there exists a policy $\coll\policy\in\ps(\coll\mdp)$  satisfying
\begin{align*}
    \precc{\coll\mdp}{\coll\policy}{\{\bot_F,\bot_{F\land F'}\}}&\geq\precc\mdp\policy F\\
    \precc{\coll\mdp}{\coll\policy}{\{\bot_{F'},\bot_{F\land F'}\}}&\geq\precc\mdp\policy {F'}\geq\thresh
\end{align*}
By \cref{prop:policyreach},
 there exists a mixture $\policy'\in\pmd(\coll\mdp)$ of two deterministic stationary policies satisfying
\begin{align*}
    \precc{\coll\mdp}{\policy'}{\{\bot_F,\bot_{F\land F'}\}}&\geq\precc{\coll\mdp}{\coll\policy}{\{\bot_F,\bot_{F\land F'}\}}\\
    \precc{\coll\mdp}{\policy'}{\{\bot_{F'},\bot_{F\land F'}\}}&\geq\thresh
\end{align*}
Moreover, by \cref{lem:abs1},
\begin{align*}
    \precc{\mdp}{\lift\policy'} F&= \precc{\coll\mdp}{\policy'}{\{\bot_F,\bot_{F\land F'}\}}\\
    \precc{\mdp}{\lift\policy'}{F'}&=\precc{\coll\mdp}{\policy'}{\{\bot_{F'},\bot_{F\land F'}\}}
\end{align*}
Hence,
\begin{align*}
    \precc{\mdp}{\lift\policy'} F&\geq \precc{\mdp}{\policy} F \\
    \precc{\mdp}{\lift\policy'}{F'}&\geq\thresh
\end{align*}
The claim follows since $\policy\in\pmix(\mdp)$ was arbitrary and $\lift\policy'\in\pmd(\mdp)$.

Next, for the second part, suppose $\policy\in\pmix(\mdpaug)$ satisfies
\begin{align*}
\precc\mdp\policy{F'\times\{0,1\}}\geq\thresh
\end{align*}
Recall that by \cref{def:mdpaug} the augmentation-bit is updated deterministically. Hence, there exists $\policy_0\in\pmix(\mdp)$ satisfying 
\begin{align*}
\precc\mdp{\policy_0}{F}&=\precc\mdp\policy{F\times\{0,1\}}\\
\precc\mdp{\policy_0}{F'}&=\precc\mdp\policy{F'\times\{0,1\}}\geq\thresh
\end{align*}
The argument proceeds as above, invoking the determinisic-lifting part of \cref{lem:abs1} to yield
\begin{align*}
    \precc{\mdpaug}{\liftdet{(\policy')}} {F\times\{0,1\}}&\geq \precc{\mdpaug}{\policy} {F\times\{0,1\}} \\
    \precc{\mdpaug}{\liftdet{(\policy')}}{F'\times\{0,1\}}&\geq\thresh
\end{align*}

\end{proof}

\section{Supplementary Materials for \cref{sec:rl}}

\transientcoll*
\begin{proof}
Let $\coll\mdp=(\coll S,\coll A,\coll\act,\coll s_0,\coll P)\defeq\collmdp{\mdp}{F}{F'}$ and
fix an arbitrary
$\policy\in\pd(\coll\mdp)$.
We construct a deterministic stationary policy
$\policy'\in\pd(\mdp)$
by lifting $\policy$ similarly to the proof of
\cref{lem:abs1}: whenever $\policy$ prescribes leaving a MEC
$\ec$ via an exit action $(s,a)\in\out(\ec)$,
$\policy'$ follows deterministically a shortest path in $\ec$ towards $s$ and then
executes $a$; whenever $\policy$ prescribes $\stay$,
$\policy'$ deterministically selects arbitrary actions in $\ec$ forever.

Define the embedded process
$(\run_i^{\coll S})_{i\ge0}$
associated with $\run\sim\distrthree{\policy'}\mdp$ as in \cref{lem:abs1}.
Analogously, to \cref{eq:markov} this process is a Markov chain with transition kernel
$\coll P^\policy$. Hence,
\begin{align*}
    B_{\mdp}\geq\E_{\run\sim\distrtwo{\policy'}}\left[\tau_{\rho,R_{\policy'}}\right]\geq \E_{\run\sim\distrtwo{\policy'}}\left[\sum_{t\geq 0}\ind{\policy(\ec_{\runstate_t})\neq\stay}\right]\geq 
    \E_{\run\sim\distrtwo{\policy'}}\left[\sum_{t\geq 0}\ind{\policy(\run^{\coll S}_t)\neq\stay}\right]
    =
    \E_{\coll\run\sim\distrthree{\policy}{\coll\mdp}}\left[\sum_{t\geq 0}\ind{\policy(\coll\run^{\coll S}_t)\neq\stay}\right]
    =\E_{\coll\run\sim\distrthree{\policy}{\coll\mdp}}\left[\tau_{\coll\run,R_\policy}-1\right]
\end{align*}
since $R_\policy=\{\bot,\bot_F,\bot_{F'},\bot_{F\land F'}\}$ and a state $s\in S$ can only be recurrent under $\policy'$ if $\policy(\ec_s)=\stay$ for the unique MEC $\ec_s$ containing $s$. \dw{justify}
The claim follows since $\policy\in\pd(\coll\mdp)$ was arbitrary.
\end{proof}

\recsample*
\begin{proof}
    Let $\mdp$ be an MDP $(S,A,\act,s_0,P)$ satisfying $B_\mdp\leq B$ and $p_\mdp\geq p$, and let $F,F'\subseteq S$, $\thresh\in[0,1]$ and $\epsilon\in(0,1)$. W.l.o.g.\ $B\geq 1$.

Let $\coll\mdp=(\coll S,\coll A,\coll\act,\coll s_0,\coll P)\defeq\collmdp\mdp F{F'}$ and suppose \pref{crec} is feasible, \changed[dw]{with $\policy'\in\pmix(\mdp)$ satisfying $\precc\mdp\policy{F'}\geq\thresh $.
By \cref{lem:abs2,lem:lptopolicy} there exists a policy $\coll\policy\in\ps(\coll\mdp)$  satisfying
\begin{align*}
    \precc{\coll\mdp}{\coll\policy}{\{\bot_F,\bot_{F\land F'}\}}&\geq\precc\mdp{\policy'} F\\
    \precc{\coll\mdp}{\coll\policy}{\{\bot_{F'},\bot_{F\land F'}\}}&\geq\precc\mdp{\policy'} {F'}\geq\thresh
\end{align*}
In particular, \pref{creach} is feasible for $\coll\mdp$, $\{\bot_F,\bot_{F\land F'}\}$, $\{\bot_{F'},\bot_{F\land F'}\}$ and $\thresh$.
}

By \cref{lem:empmecs}, with probability at least 
    \begin{align}
    \label{eq:fail1}
        1 - |S|^2|A|\cdot \exp\!\left(-N\cdot p_\mdp\right) - |S||A| \cdot\exp\!\left(-N/B_\mdp\right)
    \end{align}
    we have $\mecs(\mdp) = \mecs(S,A,\act,\widehat E)$.
Conditioned on this event, \cref{alg:rlrec} constructs the true collapsed MDP structure $(\coll S,\coll A,\coll\act,\coll s_0)$ of $\coll\mdp$ and the samples $\coll{\mathcal D}$ are distributed according to $\coll P$. 
By \cref{lem:transientcoll}, $B+1\geq B_{\mdp}+1\geq B_{\coll\mdp}$ and therefore by \cref{prop:reachrl}, with probability at least
\begin{align}
\label{eq:fail2}
    1 - c\cdot|S||A|\cdot\exp\!\left(-c'\cdot\frac{N\epsilon^2}{B+1}\right)
\end{align}
$\policy\in\pmd(\coll\mdp)$ in line~\ref{alg:creachline} of \cref{alg:rlrec} is an $\epsilon$-optimal solution of \pref{creach} for $\coll\mdp$, $\{\bot_F,\bot_{F\land F'}\}$, $\{\bot_{F'},\bot_{F\land F'}\}$ and $\thresh$.
\changed[dw]{In particular,
\begin{align*}
    \precc{\coll\mdp}{\policy}{\{\bot_F,\bot_{F\land F'}\}}&\geq \precc{\coll\mdp}{\coll\policy}{\{\bot_F,\bot_{F\land F'}\}}-\epsilon\\
    \precc{\coll\mdp}{\policy}{\{\bot_{F'},\bot_{F\land F'}\}}&\geq\thresh-\epsilon
\end{align*}
Moreover, by \cref{lem:abs1},
\begin{align*}
    \precc{\mdp}{\lift\policy} F&= \precc{\coll\mdp}{\policy}{\{\bot_F,\bot_{F\land F'}\}}\\
    \precc{\mdp}{\lift\policy}{F'}&=\precc{\coll\mdp}{\policy}{\{\bot_{F'},\bot_{F\land F'}\}}
\end{align*}
(lifting applied to each component of the mixture independently).
Hence, 
\begin{align*}
    \precc{\mdp}{\lift\policy} F&\geq \precc{\mdp}{\policy'} F-\epsilon\\
    \precc{\mdp}{\lift\policy}{F'}&\geq\thresh-\epsilon
\end{align*}
Since $\policy'$ was an arbitrary feasible policy, it follows that $\lift\policy$ is $\epsilon$-optimal.
}

Finally, we observe that, by union bound over the two failure probabilities of \cref{eq:fail1,eq:fail2}, with probability at least
\begin{align*}
    &1 - |S|^2|A|\cdot \exp\!\left(-N\cdot p_\mdp\right) - |S||A| \cdot\exp\!\left(-N/B_\mdp\right)-c\cdot|S||A|\cdot\exp\!\left(-c'\cdot\frac{N\epsilon^2}{B+1}\right)\\
    &\geq 1 - |S|^2|A|\cdot \exp\!\left(-N\cdot p\right) - (c+1)\cdot|S||A| \cdot\exp\!\left(-\min\left\{1,\frac {c'}2\right\}\cdot\frac{ N\epsilon^2} B\right)
\end{align*}
the returned $\lift\policy$ is an $\epsilon$-optimal solution of \pref{crec} for $\mdp$, $F$, $F'$ and $\thresh$.
\end{proof}

\asconv*
\dw{The prose exposition forgot to mention $p_N\defeq N^{-1/4}$, which isn't needed when describing the method but when invoking \cref{thm:recsample}}
\begin{proof}
Recall that $B_N\defeq N^{1/4}$ and $\epsilon_N\defeq N^{-1/4}$. \changed[dw]{We also define $p_N\defeq N^{-1/4}$.}

First, we fix $\epsilon>0$.
Clearly there exists $N_0\geq 1$ such that $B_N\geq B_{\mdp}$, $p_\mdp\geq p_N$ and $\epsilon_N\leq\epsilon$ for all $N\geq N_0$.
By \cref{thm:recsample} (invoked with $B_N$, $P_N$ and $\epsilon_N$), with probability at least $1-\delta_N$, $\policy_N$ is $\epsilon_N$-optimal (hence $\epsilon$-optimal), where
\begin{align*}
    \delta_N\defeq 
     c\cdot|S||A| \cdot\exp\!\left(-c'\cdot\frac{ N\epsilon_N^2} {B_N}\right)+|S|^2|A|\cdot \exp\!\left(-N\cdot p_N\right)
\end{align*}
Note that
\begin{align*}
    \sum_{N\geq 1}\delta_N&= c\cdot|S||A| \cdot\left(\sum_{N\geq 1}\exp\!\left(-c'\cdot\frac{ N\epsilon_N^2} {B_N}\right)\right)+|S|^2|A|\cdot \sum_{N\geq 1}\exp\!\left(-N\cdot p_N\right)\\
    &=c\cdot|S||A| \cdot\left(\sum_{N\geq 1}\exp\!\left(-c'\cdot N^{\frac 1 4} \right)\right)+|S|^2|A|\cdot \sum_{N\geq 1}\exp\!\left(-N^{\frac 3 4}\right)\\
    &<\infty
\end{align*}
Hence, by the first Borel-Cantelli lemma, the event
\begin{quote}
    $\policy_N$ is \emph{not} $\epsilon$-optimal for infinitely many $N\geq 1$
\end{quote}
has probability 0, and hence the event $\mathcal E_{\epsilon}$ defined as
\begin{quote}
    there exists $N_1\geq 1$ such that $\policy_N$ is $\epsilon$-optimal for all $N\geq N_1$
\end{quote}
occurs almost surely.
Intersecting the (countable) sequence of probability-one events $\mathcal E_{\epsilon_N}$ yields the claim since $\epsilon_N\to 0$.
\end{proof}

\subsection{Proof of \cref{lem:empmecs}}
Fix an $\mdp = (S,A,\act,s_0,P)$.
\empmecs*
 We first generalise the definition of end component to arbitrary relations $E \subseteq S \times A \times S$.
\begin{definition}
    For each $E \subseteq S \times A \times S$, an \emph{end component} (EC) of $(S,A,\act,E)$ is a pair
    $(S_\ec,\act_\ec)$, where $S_\ec\subseteq S$ and $\act_\ec:S_\ec\to 2^A$, satisfying:
    \begingroup
\renewcommand{\theenumi}{\arabic{enumi}}
\renewcommand{\labelenumi}{\theenumi.}
\begin{enumerate}
    \item $\act_\ec(s)\subseteq\act(s)$ for every $s\in S_\ec$,
    \item If $s \in S_\ec$, $a \in \act_\ec(s)$, and $(s,a,s') \in E$ then $s' \in S_\ec$,
    \item the graph $\left( S_\ec,\rightarrow^E_{\act_\ec} \right)$ is strongly connected,
    where $s\rightarrow^E_{\act_\ec}s'$ iff
    $(s,a,s') \in E$ for some $a\in\act_\ec(s)$.
\end{enumerate}
\endgroup
\end{definition}
As before, we let $\mecs(S,A,\act,E)$ denote the set of maximal end components of $(S,A,\act,E)$ under the containment partial order. Note in particular that $\mecs(\mdp) = \mecs(S,A,\act,\supp(P))$.

\begin{lemma}
    \label{lem:empmecsinternal}
    If $E \subseteq \supp(P)$ contains every positive-probability internal transition of $\mdp$, then each $\ec \in \mecs(\mdp)$ is contained in some $\ec' \in \mecs(S,A,\act,E)$.
\end{lemma}
\begin{proof}
    Let $\ec \in \mecs(\mdp)$. It suffices to show that $\ec$ is an EC of $(S,A,\act,E)$. We verify the three requirements:
    \begingroup
\renewcommand{\theenumi}{\arabic{enumi}}
\renewcommand{\labelenumi}{\theenumi.}
    \begin{enumerate}
        \item It is automatic from $\ec \in \mecs(\mdp)$ that $\act_\ec(s) \subseteq \act(s)$ for every $s \in S_\ec$.
        \item Suppose $s \in S_\ec$, $a \in \act_\ec(s)$, and $s' \in S$. We confirm
        \begin{align*}
            (s,a,s') \in E \Longrightarrow (s,a,s') \in \supp(P) \Longrightarrow s' \in S_\ec
        \end{align*}
        \item By assumption, the graph $\left(S_\ec, \to^{\supp(P)}_{\act_\ec}\right)$ is strongly-connected. Letting $s,s' \in S_\ec$, we observe
        \begin{align*}
            s \to^{\supp(P)}_{\act_\ec} s' &\Longrightarrow (s,a,s') \in \supp(P) \text{ for some } a \in \act_\ec(s) \\
            &\Longrightarrow (s,a,s') \in E \text{ for some } a \in \act_\ec(s) \\
            &\Longrightarrow s \to^E_{\act_\ec} s'
        \end{align*}
        as $E$ contains all positive-probability internal transitions of $\mdp$. Hence $\left( S_\ec, \to^E_{\act_C} \right)$ is also strongly connected.
    \end{enumerate}
    \endgroup
\end{proof}

\begin{definition}
    For $X \subseteq \{ (s,a) \mid s \in S, a \in \act(s) \}$, we define the directed graph $G(X) \defeq (S,\rightarrow_X)$ where
    \begin{align*}
        s \to_X s' \text{ iff there is } a \in \act(s) \text{ with } (s,a) \in X \text{ and }  P(s' \mid s,a) > 0
    \end{align*}
    Furthermore, we use $\text{SCC}_X(s) \subseteq S$ to denote the strongly connected component of $G(X)$ containing $s \in S$.
\end{definition}

\lw{should this be a subsection?}

We now construct finite sequences $(X_i)_{i \in [k+1]}, (s_i,a_i)_{i\in[k]}$ for some $0 \leq k \leq |S||A|$ which will be referenced in subsequent lemmas. Let $X_1 \defeq \{ (s,a) \mid s \in S, a \in \act(s) \}$.
    We recursively construct a \emph{strictly} descending sequence
    \begin{align*}
        X_1 \supset X_2 \supset \dots \supset X_{k+1}
    \end{align*}
    Supposing $X_i$ has been constructed, we terminate if
    \begin{align*}
        \sum_{s' \in \text{SCC}_{X_i}(s)} P(s' \mid s,a) = 1
    \end{align*}
    for all $(s,a) \in X_i$. Otherwise, there exists $(s_i,a_i) \in X_i$ with $\sum_{s' \in S \setminus \text{SCC}_{X_i}(s_i)} P(s' \mid s_i,a_i) > 0$. Without loss of generality, we may assume $(s_i,a_i)$ exits $\text{SCC}_{X_i}(s_i)$ with maximal probability:
    \begin{align*}
        (s_i,a_i) \in \underset{(s,a) \in X_i \text{ with } s \in \text{SCC}_{X_i}(s_i)}{\arg\max} \ \sum_{s' \in S \setminus \text{SCC}_{X_i}(s_i)} P(s' \mid s,a)
    \end{align*}
    We then choose $X_{i+1} \defeq X_i \setminus \{ (s_i,a_i) \}$. As a distinct state-action pair is removed each iteration, we must have $k \leq |S||A|$.

\begin{definition}
   We let $X_{\text{EC}}$ denote the set of state-action pairs contained in some $\ec \in \mecs(\mdp)$:
    \begin{align*}
        X_{\text{EC}} \defeq \{ (s,a) \mid \text{There exists } \ec \in \mecs(\mdp) \text{ with } s \in S_\ec \text{ and } a \in \act_\ec(s) \}
    \end{align*}
\end{definition}

\begin{lemma}
    \label{lem:trueecactions}
    $X_{\mathrm{EC}} = X_{k+1}$
\end{lemma}
\begin{proof}
    $(X_{\text{EC}} \subseteq X_{k+1})$ We inductively show $X_{\text{EC}} \subseteq X_i$ for $i \in [k+1]$. The base case $i = 1$ is immediate. Now suppose $X_{\text{EC}} \subseteq X_i$ for some $i \in        [k]$. Recall that $(s_i,a_i)$ satisfies
    \begin{align*}
        \sum_{s' \in S \setminus \text{SCC}_{X_i}(s_i)} P(s' \mid s_i,a_i) > 0
    \end{align*}
    That is, $(s_i,a_i)$ has positive probability of transitioning to a state $s' \in S$ from which there is no path in $G(X_i)$ back to $s_i$. By the inductive hypothesis, there is no path from $s'$ to $s_i$ in $G(X_{\text{EC}})$ either. Then $s_i,s'$ must belong to distinct MECs of $\mdp$, and $(s_i,a_i) \notin X_{\text{EC}}$. Thus $X_{i+1} = X_i \setminus \{(s_i,a_i)\} \supseteq X_{\text{EC}}$ as needed.

    $(X_{\text{EC}} \supseteq X_{k+1})$ Let $(s^*,a^*) \in X_{k+1}$, and define $\ec \defeq (S_\ec, \act_\ec)$ by
    \begin{align*}
        S_\ec &\defeq \text{SCC}_{X_{k+1}}(s^*) \\
        \act_\ec(s) &\defeq \{ a \in \act(s) \mid (s,a) \in X_{k+1} \} && s \in \text{SCC}_{X_{k+1}}(s^*)
    \end{align*}
    It suffices to show that $\ec$ is an EC of $\mdp$.
    First, suppose $s \in \text{SCC}_{X_{k+1}}(s^*)$ and $a \in \act(s)$ with $(s,a) \in X_{k+1}$. The termination condition yields
    \begin{align*}
        \sum_{s \in \text{SCC}_{X_{k+1}}(s^*)} P(s' \mid s,a) = \sum_{s' \in \text{SCC}_{X_{k+1}}(s)} P(s' \mid s,a) = 1
    \end{align*}
    where we have used $\text{SCC}_{X_{k+1}}(s) = \text{SCC}_{X_{k+1}}(s^*)$. Next, observe that
    \begin{align*}
        s \to_{X_{k+1}} s' &\Longrightarrow P(s' \mid s,a) > 0 \text{ for some } a \in \act(s) \text{ with } (s,a) \in X_{k+1} \\
        &\Longrightarrow (s,a,s') \in \supp(P) \text{ for some } a \in \act_\ec(s) \\
        &\Longrightarrow s \to^{\supp(P)}_{\act_\ec} s'
    \end{align*}
    for $s,s' \in \text{SCC}_{X_{k+1}}(s^*)$. Hence $\left( \text{SCC}_{X_{k+1}}(s^*), \to^{\supp(P)}_{\act_\ec} \right)$ inherits strong connectivity from $\left( \text{SCC}_{X_{k+1}}(s^*), \to_{X_{k+1}} \right)$ as needed.
\end{proof}

\begin{lemma}
    \label{lem:elimcontain}
    If $E \subseteq \supp(P)$ and $\ec \in \mecs(S,A,\act,E)$ satisfies
    \begin{align*}
        \text{for all } s \in S_\ec \text{ and } a \in \act_\ec(s), \ (s,a) \in X_{\mathrm{EC}}
    \end{align*}
    then $\ec$ is contained in some $\ec' \in \mecs(\mdp)$.
\end{lemma}
\begin{proof}
    Let $\ec \in \mecs(S,A,\act,E)$ be as stated. Fix $s^* \in S_\ec$, and let $\ec^* \in \mecs(\mdp)$ be the unique MEC of $\mdp$ containing $s^*$. We claim $\ec$ is contained in $\ec^*$. Due to the assumption on $\ec$, it suffices to verify $S_\ec \subseteq S_{\ec^*}$.

    For each $s \in S_\ec$, strong connectivity of $\ec$ guarantees a path from $s^*$ to $s$ in $\left( S_\ec, \to^E_{\act_\ec} \right)$. We will show $s \in S_{\ec^*}$ by induction on path length $n$. The base case $n = 0$ is immediate. Now consider the final edge $s' \to^E_{\act_\ec} s$ on a path of length $n+1$, where the inductive hypothesis yields $s' \in S_{\ec^*}$. By the definition of edge, there exists $a \in \act_\ec(s')$ with $(s',a,s) \in E \subseteq \supp(P)$. The assumption on $\ec$ gives $(s',a) \in X_{\text{EC}}$. Thus $a \in \act_{\ec^*}(s')$, and the closure property of $\ec^*$ gives $s \in S_{\ec^*}$ as claimed.
\end{proof}

\lw{Use $\max(B_\mdp, 1)$ to avoid division by zero}
\begin{lemma}
    \label{lem:empmecslemma}
    If $E \subseteq \supp(P)$ satisfies
    \begin{align*}
        \text{for all } i \in [k], \text{ there exists } s' \in S \setminus \mathrm{SCC}_{X_i}(s_i) \text{ with } (s_i,a_i,s') \in E
    \end{align*}
    then each $\ec \in \mecs(S,A,\act,E)$ is contained in some $\ec' \in \mecs(\mdp)$.
\end{lemma}
\begin{proof}
    Let $\ec \in \mecs(S,A,\act,E)$. By \cref{lem:elimcontain}, it suffices to show that $(s,a) \in X_{\text{EC}}$ for all $s \in S_\ec$ and $a \in \act_\ec(s)$. If this were not the case, there would be $i \in [k]$ with $s_i \in S_\ec$ and $a_i \in \act_\ec(s_i)$ by \cref{lem:trueecactions}. Hence we may suppose towards a contradiction that $i \in [k]$ is minimal such that $s_i \in S_\ec$ and $a_i \in \act_C(s)$.
    
    By assumption, there exists $s' \in S \setminus \text{SCC}_{X_i}(s_i)$ with $(s_i,a_i,s') \in E$. The closure property of $\ec$ yields $s' \in S_\ec$, and strong connectivity of $\left( S_\ec, \to^E_{\act_\ec} \right)$ guarantees a path from $s'$ back to $s_i$. Each edge $s \to^E_{\act_\ec} t$ on this path corresponds to some $a \in \act_\ec(s)$ with $(s,a,t) \in E \subseteq \supp(P)$. By minimality of $i$, $(s,a) \notin \{ (s_j,a_j) \mid j \in [i-1] \}$. Equivalently, $(s,a) \in X_i$. Then $s \to_{X_i} t$ for each such edge, and the full path from $s'$ to $s_i$ is present in $G(X_i)$. Likewise, since $(s_i,a_i) \in X_i$ and $(s_i,a_i,s') \in E \subseteq \supp(P)$, we have $s_i \to_{X_i} s
'$. Thus $s_i$ and $s'$ are mutually reachable in $G(X_i)$, contradicting $s' \in S \setminus \text{SCC}_{X_i}(s_i)$.
\end{proof}

\begin{lemma}
    \label{lem:empmecsbprob}
    For $i \in [k]$, we have
    \begin{align*}
        \sum_{s' \in S \setminus \mathrm{SCC}_{X_i}(s_i)} P(s' \mid s_i,a_i) \geq \frac{1}{B_\mdp}
    \end{align*}
\end{lemma}
\begin{proof}
    As $\text{SCC}_{X_i}(s_i)$ is strongly connected, we can construct a stationary deterministic policy $\policy$ with $\policy(s_i) = a_i$, and at all other states in $\text{SCC}_{X_i}(s_i)$ performs an action corresponding to the first step of a $G(X_i)$-path to $s_i$. By construction, $\policy$ has positive probability of reaching $s_i$ when starting from any state in $\text{SCC}_{X_i}(s_i)$. Hence no state in $\text{SCC}_{X_i}(s_i)$ can be recurrent for $\policy$, as this would imply $(s_i,a_i) \in X_{\text{EC}}$ by \cref{lem:alfaro}.
    
    Now fix an $s^* \in \text{SCC}_{X_i}(s_i)$ minimizing the expected hitting time of a recurrent state under $\policy$.
    \begin{align*}
        s^* \in \arg\min_{\text{SCC}_{X_i}(s_i)} \E_{\run\sim\distrthree{\policy}{\mdp[s^*]}}[\tau_{\run,R_{\policy}}]
    \end{align*}
    Minimality, in combination with $R_{\policy} \cap \text{SCC}_{X_i}(s_i) = \emptyset$, gives
    \begin{align*}
        \E_{\run\sim\distrthree{\policy}{\mdp[s^*]}}[\tau_{\run,R_{\policy}}] \geq 1 + \left( \sum_{s' \in \text{SCC}_{X_i}(s_i)} P(s' \mid s^*, \policy(s^*)) \right) \cdot \E_{\run\sim\distrthree{\policy}{\mdp[s^*]}}[\tau_{\run,R_{\policy}}]
    \end{align*}
    Hence we have
    \begin{align*}
        \sum_{s' \in S \setminus \text{SCC}_{X_i}(s_i)} P(s' \mid s_i, a_i) \geq \sum_{s' \in S \setminus \text{SCC}_{X_i}(s_i)} P(s' \mid s^*, \policy(s^*)) \geq \frac{1}{\E_{\run\sim\distrthree{\policy}{\mdp[s^*]}}[\tau_{\run,R_{\policy}}]} \geq \frac{1}{B_\mdp}
    \end{align*}
    as $(s_i,a_i)$ was chosen to maximise the probability of leaving $\text{SCC}_{X_i}(s_i)$.
\end{proof}

\empmecs*
\begin{proof}
    The probability of not observing an individual positive-probability internal transition of $\mdp$ is at most
    \begin{align*}
        (1-p_\mdp)^N \leq e^{-N \cdot p_\mdp}
    \end{align*}
    By a naive union bound, the event ``$\widehat E$ contains all positive-probability internal transitions of $\mdp$'' has probability at least
    \begin{align*}
        1 - |S|^2|A| \cdot e^{-N \cdot p_\mdp}
    \end{align*}
    On this event, every true MEC is contained in an empirical MEC by \cref{lem:empmecsinternal}.

    Similarly, for each $i \in [k]$, the probability that 
    \begin{align*}
        (s_i,a_i,s') \notin \widehat E \text{ for all } s' \in S \setminus \text{SCC}_{X_i}(s_i)
    \end{align*}
    is at most
    \begin{align*}
        \left( 1 - \frac{1}{B_\mdp} \right)^N \leq e^{-N/B_\mdp}
    \end{align*}
    by \cref{lem:empmecsbprob}. Since $k \leq |S||A|$, the event
    \begin{align*}
        \text{for all } i \in [k], \text{ there exists } s' \in S \setminus \text{SCC}_{X_i}(s_i) \text{ with } (s_i,a_i,s') \in \widehat E
    \end{align*}
    has probability at least
    \begin{align*}
        1 - |S||A| \cdot e^{-N/B_\mdp}
    \end{align*}
    On this event, every empirical MEC is contained in a true MEC by \cref{lem:empmecslemma}.

    Thus both events hold simultaneously with probability at least
    \begin{align*}
        1 - |S|^2|A|\cdot e^{-N\cdot p} - |S||A| \cdot e^{-N/B_\mdp}
    \end{align*}
    on which the bidirectional containments give $\mecs(\mdp) = \mecs(S,A,\act,\widehat E)$.
\end{proof}

\section{Supplementary Materials for \cref{sec:lower}}

\begin{theorem}
\label{thm:lowerboundb}
    For every $n \in \nat$, $m \geq 4$, $B \geq 1$, and $\epsilon \in (0,1/4)$, there exists a family of unconstrained reachability problems $(\mdp, F)$ such that:
\begingroup
\renewcommand{\theenumi}{\arabic{enumi}}
\renewcommand{\labelenumi}{\theenumi.}
    \begin{enumerate}
        \item every $\mdp = (S,A,\act,s_0,P)$ satisfies $|S| = 6n$, $|A| = m + 1$, $B_\mdp = B$, and $p_\mdp = 1$;
        \item any generative-model RL algorithm that returns an $\epsilon$-optimal policy with probability at least $3/4$ on every problem in the family has expected sample complexity $\Omega\left(B / \epsilon^2 \right)$ per state-action pair on at least one instance.
    \end{enumerate}
\endgroup
\end{theorem}
\begin{proof}
    We adapt \cite[Thm.~4]{zurek2024span} to the reachability setting. They form MDPs by composing $4$-state gadgets, then show that the resulting family of unconstrained average-reward problems require $\Omega(B/\epsilon^2)$ generative model queries per state-action pair. \cref{fig:reachability-comparison} gives a conversion from their $4$-state reward gadgets to $3$-state reachability gadgets. The absorbing state granting reward $1/2$ is simulated using one absorbing target state and one absorbing non-target state, each reached with equal probability via the dedicated action $a_0$. It is readily seen that reachability satisfaction in the new gadget corresponds exactly to average reward in the original gadget. As the transformation preserves $|S|$, $|A|$, and $B$ asymptotically, unconstrained reachability inherits the $\Omega(B/\epsilon^2)$ sample complexity, per state-action pair, of unconstrained average reward. Finally, we remark that all positive-probability internal transitions in each gadget have probability $1$, hence $p_\mdp = 1$ for each constructed MDP $\mdp$.
\end{proof}

\begin{table}[t]
\centering
\setlength{\tabcolsep}{6pt}
\renewcommand{\arraystretch}{1.15}

\begin{tabular}{m{4cm}|m{0.36\linewidth}|m{0.4\linewidth}}
& Null gadget & Positive gadget ($a^* = a_i$ for some $i \in [m]$) \\ \hline &&\\

Reachability $(F = \{s_1\})$
&
\begin{subfigure}{0.34\textwidth}
    \centering
    \scalebox{0.7}{
        \begin{tikzpicture}[node distance = 2cm]
            \node[state] (s0) {$s_0$};
            \node[state, accepting, above right of=s0] (s1) {$s_1$};
            \node[state, below right of=s0] (s2) {$s_2$};
              \draw[->] 
              (s0) edge [loop left] node [left]{$\{a_1,\dots,a_m\}/1-\frac{1}{B}$} (s0)
                (s0) edge node [above left]{$a_0/\frac{1}{2}, \{a_1,\dots,a_m\} / \frac{1-2\epsilon}{2B}$} (s1)
                (s1) edge [loop right] node [right]{$A/1$} (s1)
                (s0) edge node [below left]{$a_0/\frac{1}{2}, \{a_1,\dots,a_m\} / \frac{1+2\epsilon}{2B}$} (s2)
                (s2) edge [loop right] node [right]{$A/1$} (s2);
        \end{tikzpicture}    
    }
\end{subfigure}
&
\begin{subfigure}{0.34\textwidth}
    \centering
    \scalebox{0.7}{
        \begin{tikzpicture}[node distance = 2cm]
            \node[state] (s0) {$s_0$};
            \node[state, accepting, above right of=s0] (s1) {$s_1$};
            \node[state, below right of=s0] (s2) {$s_2$};
              \draw[->] 
              (s0) edge [loop left] node [left]{$\{a_1,\dots,a_m\}/1-\frac{1}{B}$} (s0)
                (s0) edge node [above left]{$a_0/\frac{1}{2}, a^* / \frac{1+2\epsilon}{2B}, (\{a_1,\dots,a_m\} \setminus a^*) / \frac{1-2\epsilon}{2B}$} (s1)
                (s1) edge [loop right] node [right]{$A/1$} (s1)
                (s0) edge node [below left]{$a_0/\frac{1}{2}, a^* / \frac{1-2\epsilon}{2B}, (\{a_1,\dots,a_m\} \setminus a^*) / \frac{1+2\epsilon}{2B}$} (s2)
                (s2) edge [loop right] node [right]{$A/1$} (s2);
        \end{tikzpicture}
    }
\end{subfigure}
\\
\hline &&\\

Avg. Reward \citep[Thm.~4]{zurek2024span}
&
\begin{subfigure}{0.34\textwidth}
    \centering
    \scalebox{0.7}{
    \begin{tikzpicture}[node distance=2cm]
            \node[state] (s0) {$s_0$};
            \node[state, above right of=s0] (s1) {$s_1$};
            \node[state, right of=s0] (s2) {$s_2$};
            \node[state, below right of=s0] (s3) {$s_3$};
              \draw[->] 
              (s0) edge [loop left] node [left]{$\{a_1,\dots,a_m\}/1-\frac{1}{B}$} (s0)
                (s0) edge node [above left]{$\{a_1,\dots,a_m\} / \frac{1-2\epsilon}{2B}$} (s1)
                (s1) edge [loop right] node [right]{$A/1, R = 1$} (s1)
                (s0) edge node [above]{$a_0/1$} (s2)
                (s2) edge [loop right] node [right]{$A/1, R=1/2$} (s2)
                (s0) edge node [below left]{$\{a_1,\dots,a_m\} / \frac{1+2\epsilon}{2B}$} (s3)
                (s3) edge [loop right] node [right]{$A/1, R = 0$} (s3);
    \end{tikzpicture}
    }
\end{subfigure}
&
\begin{subfigure}{0.34\textwidth}
    \centering
    \scalebox{0.7}{
    \begin{tikzpicture}[node distance=2cm]
            \node[state] (s0) {$s_0$};
            \node[state, above right of=s0] (s1) {$s_1$};
            \node[state, right of=s0] (s2) {$s_2$};
            \node[state, below right of=s0] (s3) {$s_3$};
              \draw[->] 
              (s0) edge [loop left] node [left]{$\{a_1,\dots,a_m\}/1-\frac{1}{B}$} (s0)
                (s0) edge node [above left]{$a^* / \frac{1+2\epsilon}{2B}, (\{a_1,\dots,a_m\} \setminus a^*) / \frac{1-2\epsilon}{2B}$} (s1)
                (s1) edge [loop right] node [right]{$A/1, R = 1$} (s1)
                (s0) edge node [above]{$a_0/1$} (s2)
                (s2) edge [loop right] node [right]{$A/1, R=1/2$} (s2)
                (s0) edge node [below left]{$a^* / \frac{1-2\epsilon}{2B}, (\{a_1,\dots,a_m\} \setminus a^*) / \frac{1+2\epsilon}{2B}$} (s3)
                (s3) edge [loop right] node [right]{$A/1, R = 0$} (s3);
    \end{tikzpicture}
    }
\end{subfigure}
\end{tabular}

\caption{Comparison of gadgets for unconstrained reachability and unconstrained average reward. }
\label{fig:reachability-comparison}
\end{table}

\begin{figure}
    \begin{center}
    \begin{subfigure}{0.55\linewidth}
    \begin{center}
    \scalebox{0.7}{
            \begin{tikzpicture}[node distance = 2cm]
            \node[state] (si) {$s_i$};
            \node[state, below left of=si] (u0) {$u_0$};
            \node[state, accepting, left of=u0] (t0) {$t_0$};
            \node[state, accepting, below right of=si] (ti) {$t_i$};
            \node[state, below of=ti] (ui) {$u_i$};
              \draw[->] 
              (si) edge [bend right] node [above left]{$a_0/\frac{1}{2}$} (t0)
                (t0) edge [loop below] node [below]{$A/1$} (t0)
                (si) edge node [above left]{$a_0/\frac{1}{2}$} (u0)
                (u0) edge [loop below] node [below]{$A/1$} (u0)
                (si) edge node [above right]{$\{a_1, \dots, a_m\}/1$} (ti)
                (ti) edge node [right]{$A/1$} (ui)
                (ui) edge [loop left] node [left]{$A/1$} (ui);
        \end{tikzpicture}    
        }
        \end{center}
    \caption{Null gadget ($f(i) = 0$).}      
    \end{subfigure}
    \hfill
\begin{subfigure}{0.43\linewidth}
    \begin{center}
    \scalebox{0.7}{
        \begin{tikzpicture}[node distance = 2cm]
            \node[state] (si) {$s_i$};
            \node[state, below left of=si] (u0) {$u_0$};
            \node[state, accepting, left of=u0] (t0) {$t_0$};
            \node[state, accepting, below right of=si] (ti) {$t_i$};
            \node[state, below of=ti] (ui) {$u_i$};
              \draw[->] 
              (si) edge [bend right] node [above left]{$a_0/\frac{1}{2}$} (t0)
                (t0) edge [loop below] node [below]{$A/1$} (t0)
                (si) edge node [above left]{$a_0/\frac{1}{2}$} (u0)
                (u0) edge [loop below] node [below]{$A/1$} (u0)
                (si) edge node [above right]{$\{a_1, \dots, a_m\}/1$} (ti)
                (ti) edge [bend right] node [left]{$A/1$} (ui)
                (ui) edge [bend right] node [right]{$a_{f(i)}/p$} (ti)
                (ui) edge [loop left] node [left]{$a_{f(i)}/1-p, (A \setminus \{a_{f(i)}\})/1$} (ui);
        \end{tikzpicture}     
        }
    \end{center}
    \caption{Positive gadget ($f(i) > 0$).}
\end{subfigure}
    
    \end{center}
    \caption{Gadgets forming hard instances of unconstrained recurrence in \cref{thm:lowerboundp}.}
        \label{fig:recgadgets}
\end{figure}

\begingroup
\renewcommand{\theenumi}{\arabic{enumi}}
\renewcommand{\labelenumi}{\theenumi.}
\lowerboundp*
\endgroup

\begin{proof}
    Define
    \begin{align*}
        S &\defeq \{ s_i, t_i, u_i \mid 0 \leq i \leq n \} \\
        A &\defeq \{ a_0, \dots, a_m \}.
    \end{align*}
    For $f: [n] \to \{0\} \cup [m]$, we define the transition kernel $P_f$ by
    \begin{align*}
        P_f(s_i \mid s_0, a) &\defeq \frac{1}{n} && a \in A \\
        P_f(t_0 \mid s_i, a_0) = P_f(u_0 \mid s_i, a_0) &\defeq \frac{1}{2} \\
        P_f(t_0 \mid t_0, a) = P_f(u_0 \mid u_0, a) &\defeq 1 && a \in A \\
        P_f(t_i \mid s_i, a) &\defeq 1 && a \in A \setminus \{a_0\} \\
        P_f(u_i \mid t_i, a) &\defeq 1 && a \in A \\
        P_f(t_i \mid u_i, a_j) &\defeq \ind{j > 0, f(i) = j} \cdot p && 0 \leq j \leq m \\
        P_f(u_i \mid u_i, a_j) &\defeq 1 - \ind{j > 0, f(i) = j} \cdot p && 0 \leq j \leq m
    \end{align*}
    for $i \in [n]$. For each resulting MDP $\mdp_f \defeq (S,A,\act,s_0,P_f)$, where $\act(s) = S$, we consider the unconstrained recurrence objective of visiting the target region $F \defeq \{ t_0, \dots, t_n \}$ infinitely often. \cref{fig:recgadgets} gives an illustration.
    
    Any policy $\policy$ induces probabilities $(p^\policy_i)_{i\in[n]}$, where $p^\policy_i$ is the conditional probability of exiting $s_i$ via action $a_0$ in rollouts where $s_i$ is reached. Regardless of the value of $f(i)$, performing $a_0$ at $s_i$ achieves the recurrence objective with probability $\frac{1}{2}$. When $f(i) = 0$, performing any action other than $a_0$ at $s_i$ results in zero recurrence satisfaction. However, when $f(i) > 0$, an optimal policy $\policy^*$ performs a non-$a_0$ action at $s_i$ and achieves recurrence satisfaction of $1$ by performing $a_{f(i)}$ at $u_i$ infinitely often. Hence we have
    \begin{align*}
        \precc{\mdp_f}{\policy^*}{F} - \precc{\mdp_f}{\policy}{F} &\geq \frac{1}{2n} \cdot \sum_{i=1}^n \left((1 - p^\policy_i) \cdot \ind{f(i) = 0} + p^\policy_i \cdot \ind{f(i) > 0}\right)
    \end{align*}
    By defining
    \begin{align*}
        \ell_{i}(f,\policy) &\defeq (1 - p^\policy_i) \cdot \ind{f(i) = 0} + p^\policy_i \cdot \ind{f(i) > 0} && i \in [n] \\
        L(f,\policy) &\defeq \sum_{i=1}^n \ell_i(f,\policy),
    \end{align*}
    we note that $\precc{\mdp_f}{\policy^*}{F} - \precc{\mdp_f}{\policy}{F} \leq 1/8$ is only possible when
    \begin{align*}
        L(f,\policy) \leq \frac{n}{4}
    \end{align*}
    Now suppose $\mathcal{A}$ is an algorithm which outputs $1/8$-optimal policies with probability at least $3/4$ for the considered family of unconstrained recurrence problems. Furthermore, suppose $f$ is sampled such that
    \begin{align*}
        f(i) \sim \frac{1}{2} \cdot \delta_0 + \frac{1}{2m} \cdot \sum_{j=1}^m \delta_j
    \end{align*}
    independently for each $i \in [n]$ ($\delta_j$ is the Dirac distribution on $j$). Let $\widehat{\policy}$ be the output of $\mathcal{A}$ on the corresponding MDP $\mdp_f$. By the correctness guarantee, we must have
    \begin{align*}
        \E [L(f, \widehat{\policy})] \leq \frac{3}{4} \cdot \frac{n}{4} + \frac{1}{4} \cdot n = \frac{7n}{16}
    \end{align*}
    where we use that $L$ is bounded above by $n$.

    For $1 \leq i \leq n$ and $1 \leq j \leq m$, let $N_{i,j}$ be the number of generative model queries made for the state-action pair $(u_i, a_j)$. Observe that
    \begin{align*}
        \E \left[ p_i^{\widehat{\policy}} \mid f(i) = j \right] \geq \E \left[ p_i^{\widehat{\policy}} \cdot (1 - p)^{N_{i,j}} \mid f(i) = 0 \right] && 1 \leq i \leq n, 1 \leq j \leq m
    \end{align*}
    as queries to $(u_i, a_j)$ in the $f(i) = j$ case are indistinguishable from the $f(i) = 0$ case \dw{I think it would be good to be a bit more explicit here} when all observed transitions are self-loops. Then, by the chosen distribution of $f(i)$, we have
    \begin{align*}
        \E \left[ \ell_i(f, \widehat{\policy}) \right] &= \frac{1}{2} \cdot \E \left[ 1 - p_i^{\widehat{\policy}} \mid f(i) = 0 \right] + \frac{1}{2m} \cdot \sum_{j=1}^m \E \left[ p_i^{\widehat{\policy}} \mid f(i) = j \right] \\
        &\geq \frac{1}{2} \cdot \E \left[ 1 - p_i^{\widehat{\policy}} \mid f(i) = 0 \right] + \frac{1}{2m} \cdot \sum_{j=1}^m \E \left[ p_i^{\widehat{\policy}} \cdot (1 - p)^{N_{i,j}} \mid f(i) = 0 \right] \\
        &\geq \frac{1}{2} \cdot \E \left[ (1 - p_i^{\widehat{\policy}}) \cdot (1-p)^{N_{i,j}} \mid f(i) = 0 \right] + \frac{1}{2m} \cdot \sum_{j=1}^m \E \left[ p_i^{\widehat{\policy}} \cdot (1 - p)^{N_{i,j}} \mid f(i) = 0 \right] \\
        &= \frac{1}{2m} \cdot \sum_{j=1}^m \E \left[ (1-p)^{N_{i,j}} \mid f(i) = 0 \right] \\
        &\geq \frac{1}{2m} \cdot \sum_{j=1}^m \E \left[ 1 - p \cdot N_{i,j} \mid f(i) = 0 \right] \\
        &= \frac{1}{2} - \frac{p}{2m} \cdot \sum_{j=1}^m \E \left[ N_{i,j} \mid f(i) = 0 \right]
    \end{align*}
    The prior on $f$ yields
    \begin{align*}
        N \defeq \sum_{i=1}^n \sum_{j=1}^m \E \left[ N_{i,j} \right] \geq \frac{1}{2} \cdot \sum_{i=1}^n \sum_{j=1}^m \E \left[ N_{i,j} \mid f(i) = 0 \right],
    \end{align*}
    hence
    \begin{align*}
        \E \left[ L(f,\widehat{\policy}) \right] &= \sum_{i=1}^m \E \left[ \ell_i(f,\widehat{\policy}) \right] \\
        &\geq \sum_{i=1}^n \left( \frac{1}{2} - \frac{p}{2m} \cdot \sum_{j=1}^m \E \left[ N_{i,j} \mid f(i) = 0 \right] \right) \\
        &= \frac{n}{2} - \frac{p}{2m} \cdot \sum_{i=1}^n \sum_{j=1}^m \E \left[ N_{i,j} \mid f(i) = 0 \right] \\
        &\geq \frac{n}{2} - \frac{p}{m} \cdot N
    \end{align*}
    Combining with the earlier correctness condition yields
    \begin{align*}
        \frac{n}{2} - \frac{p}{m} \cdot N \leq \E \left[ L(f,\widehat{\policy}) \right] \leq \frac{7n}{16}
    \end{align*}
    Rearrangement then gives
    \begin{align*}
        N \geq \frac{nm}{16p} = \Omega \left( \frac{nm}{p} \right)
    \end{align*}
    Hence there must exist an $f$ in the class for which $\mathcal{A}$ performs $\Omega(nm/p)$ total generative model queries in expectation.
\end{proof}

\maincomplexity*
\begin{proof}
    \cref{thm:lowerboundb} establishes sample complexity $\Omega(B/\epsilon^2)$ for a class of MDPs satisfying $B_\mdp \leq B$ and $p_\mdp = 1$. Conversely, \cref{thm:lowerboundp} establishes sample complexity $\Omega(1/p)$ for a class of MDPs satisfying $B_\mdp \leq 3$ amd $p_\mdp \geq p$. Thus for $B \geq 3$, the class of MDPs satisfying $B_\mdp \leq B$ and $p_\mdp \geq p$ contains both identified hard classes, giving the combined bound $\Omega(1/p + B/\epsilon^2)$.
\end{proof}

\end{document}